\documentclass[conference]{IEEEtran}
\usepackage{times}

\usepackage[numbers]{natbib}
\usepackage{multicol}
\usepackage[bookmarks=true]{hyperref}
\usepackage{mathtools}
\usepackage{amssymb,amsmath}
\usepackage[mathcal]{euscript}
\usepackage{xcolor}
\usepackage{booktabs}
\usepackage[symbol]{footmisc}

\DeclarePairedDelimiter\abs{\lvert}{\rvert}

\begin{document}

\title{Simple Sensor Intentions for Exploration}


\author{\IEEEauthorblockN{Tim Hertweck\textsuperscript{*}, Martin Riedmiller\textsuperscript{*}, Michael Bloesch, Jost Tobias Springenberg,\\Noah Siegel, Markus Wulfmeier, Roland Hafner, Nicolas Heess\footnotetext[1]{Equal contribution. Correspondence to: thertweck@google.com, riedmiller@google.com.}}
\IEEEauthorblockA{DeepMind\\London, United Kingdom}}


%

\maketitle

\begin{abstract}


Modern reinforcement learning algorithms can learn solutions to increasingly difficult control problems while at the same time reduce the amount of prior knowledge needed for their application. One of the remaining challenges is the definition of reward schemes that appropriately facilitate exploration without biasing the solution in undesirable ways, and that can be implemented on real robotic systems without expensive instrumentation. In this paper we focus on a setting in which goal tasks are defined via simple sparse rewards, and exploration is facilitated via agent-internal auxiliary tasks.
We introduce the idea of simple sensor intentions (SSIs) as a generic way to define auxiliary tasks. SSIs reduce the amount of prior knowledge that is required to define suitable rewards. They can further be computed directly from raw sensor streams and thus do not require expensive and possibly brittle state estimation on real systems.
We demonstrate that a learning system based on these rewards can solve complex robotic tasks in simulation and in real world settings. In particular, we show that a real robotic arm can learn to grasp and lift and solve a Ball-in-a-Cup task from scratch, when only raw sensor streams are used
for both controller input and in the auxiliary reward definition. A video showing the results can be found at \href{https://deepmind.com/research/publications/Simple-Sensor-Intentions-for-Exploration}{https://deepmind.com/research/publications/Simple-Sensor-Intentions-for-Exploration.}
\end{abstract}

\IEEEpeerreviewmaketitle

\footnotetext[1]{Equal contribution. Correspondence to: \href{mailto:thertweck@google.com}{thertweck@google.com}, \href{mailto:riedmiller@google.com}{riedmiller@google.com}.}

\section{Introduction}

An important consideration on the path towards general AI is to minimize the amount of prior knowledge needed to set up learning systems. Ideally, we would like to identify principles 
that transfer to a wide variety of different problems without the need for manual tuning and problem-specific adjustments. 
In recent years, we have witnessed substantial progress in the ability of reinforcement learning algorithms to solve difficult control problems 
from first principles, often  directly from raw sensor signals such as camera images, without need for carefully handcrafted features that would require human understanding of the particular environment or task \cite{mnih2015humanlevel, SilverHuangEtAl16nature}.

One of the remaining challenges is the definition of reward schemes that appropriately indicate task success, facilitate exploration without biasing the solution in undesirable ways, and that can be implemented on real robotics systems without expensive instrumentation.
In this paper we focus on situations, where the external task is given by a sparse reward signal, that is `$1$' if and only if the task is solved. Such reward functions are often easy to define, and by solely focusing on task success strongly mitigate the bias on the final solution. The associated challenge is, however, that starting from scratch with a naive exploration strategy will most likely never lead to task success and the agent will hardly receive any learning signal. 

The Scheduled Auxiliary Control (SAC-X) \cite{Riedmiller2018_Learning} framework tackles this problem by introducing a set of auxiliary rewards, that help the agent to explore the environment. For each auxiliary reward an auxiliary policy (`intention') is learned and executed to collect data into a shared replay-buffer. This diverse data facilitates learning of the main task.
The important insight of \cite{Riedmiller2018_Learning} is that 
the exact definition of auxiliary tasks can vary, as long as they jointly lead to an adequate exploration strategy that allows to collect rich enough data such that learning of the main task can proceed. 
However, in the original work the auxiliary tasks are still defined with some semantic understanding of the environment in mind, e.g. `move an object' or `place objects close to each other'. This requires both task-specific knowledge for the definition of the auxiliaries, as well as the technical prerequisites to estimate the relevant quantities required for the computation of the rewards 
- e.g. camera calibration, object detection and object pose estimation. In this work we want to make a step towards a more generic approach for defining auxiliary tasks, that reduces the need for task-specific semantic interpretation of environment sensors, in particular of camera images.
%

\begin{figure}
    \centering
    \centerline{
    \includegraphics[width=.45\linewidth]{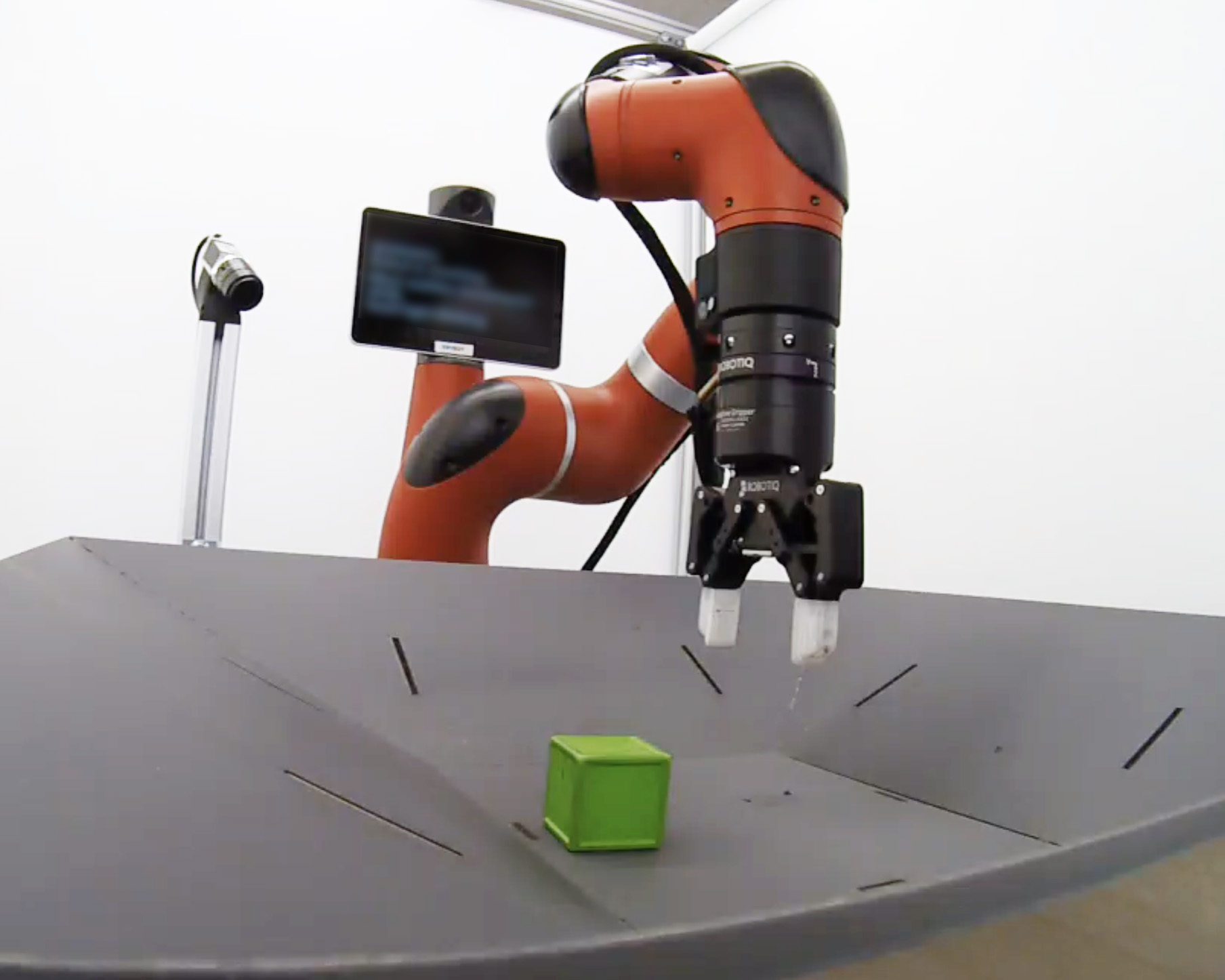} 
    \includegraphics[width=.45\linewidth]{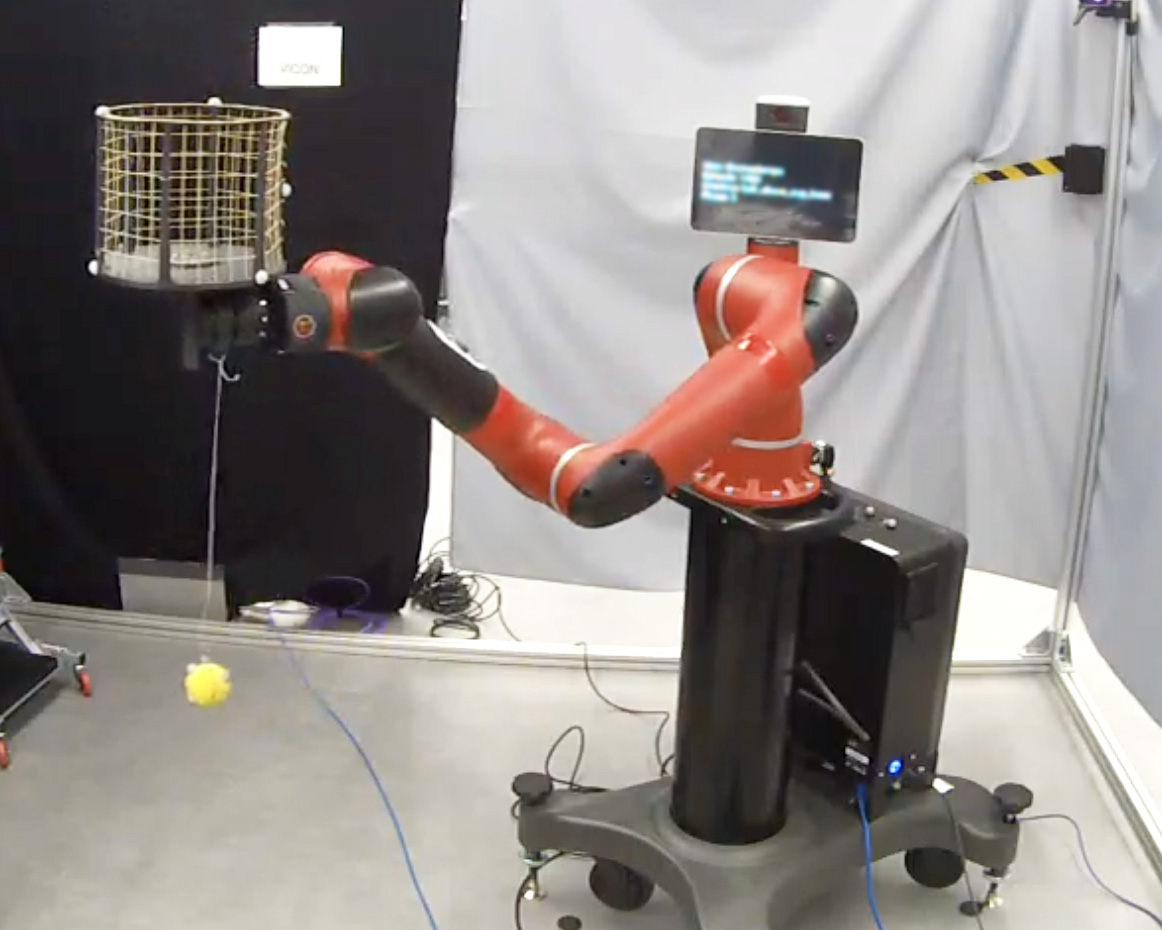}
    }
    \caption{Manipulation setup with a Rethink Sawyer robotic arm and a Robotiq 2F-85 parallel gripper (left). Ball-in-a-cup task setup with a Rethink Sawyer robotic arm and a custom made Ball-and-Cup attachment (right).}
    \label{fig:experimental_setup}
\end{figure}


A fundamental principle to enable exploration in sparse reward scenarios, is to learn auxiliary behaviours that deliberately change sensor responses. While variants of this idea have been already suggested earlier, e.g. \cite{Sutton2011horde, Jaderberg2017Unreal}, we here introduce a generic way to implement this concept into the SAC-X framework: 'simple sensor intentions' (SSIs) encourage the agent to explore the environment and help with collecting meaningful data for solving the sparsely rewarded main task. Being largely task agnostic, SSIs can be reused across tasks with no or only minimal adjustments. Further, simple sensor intentions (SSIs) are based on rewards defined on the deliberate change of scalar sensor responses, that are derived from raw sensor values. We propose two ways to effect change, namely (a) to attain certain set points (like e.g. the minimum or maximum of a sensor response), or (b) by rewarding increase or decrease of a sensor response. 

However, not all sensory inputs can be directly mapped to scalar values, like e.g. raw camera images. As an exemplary procedure, we suggest a concrete pre-processing for mapping raw images into simple sensor responses by computing and evaluating basic statistics, such as the spatial mean of color-filtered images. While SSIs propose a general way to deal with all kinds of sensor values available in a robotic system (like touch sensors, joint angle sensors, position sensors, ...), we mostly investigate pixel inputs here as an example of a sensor type that is widely used in robotics.

As a proof of concept, we show that these simple sensor intentions (SSIs) can be applied on a variety of interesting robotic domains, both in simulation and on real robots. Most notably, we show that with SSIs, SAC-X is capable of learning to play the Ball-in-a-Cup game from scratch, purely from pixel and proprioceptive inputs - for both as observation and for computing the auxiliary rewards given to the agent.

\medskip
\section{Preliminaries}

We consider a reinforcement learning setting with an agent operating in a Markov Decision Process (MDP) consisting of the state space $\mathcal{S}$, the action space $\mathcal{A}$ and the transition probability $p(s_{t+1}|s_t, a_t)$ of reaching state $s_{t+1}$ from state $s_t$ when executing action $a_t$ at the previous time step $t$. The goal of an agent in this setting is to succeed at learning a given task $k$ out of a set of possible tasks $\mathcal{K}$. 
The actions are assumed to be drawn from a probability distribution over actions $\pi_k(a|s)$ referred to as the agent's policy for task $k$. After executing an
action $a_t$ in state $s_t$ and reaching state $s_{t+1}$ the agent receives a task-dependent, scalar reward $r_k(s_t, s_{t+1})$. Given a target task $g$ we define the expected return (or value) when following the task-conditioned policy $\pi_g$, starting from state $s$, as
\begin{equation*}
V^{\pi_g}(s) = \mathbb{E}_{\pi_g} [ \: \sum_{t = 0}^{\infty} \gamma^t r_g(s_t, s_{t+1}) \: | \: s_0 = s \: ],
\end{equation*}
with $a_t \sim \pi_g(\cdot | s_t)$ and $s_{t+1} \sim p(\cdot | s_t, a_t)$ for all $s \in \mathcal{S}$.
The goal of Reinforcement Learning then is to find the policy $\pi^*_g$ that maximizes the value. The auxiliary intentions, defined in the following based on simple sensor rewards -- that is task $k \in \mathcal{K}, k \neq g$ -- give rise to their own values and policies and serve as means for efficiently exploring the MDP.

\medskip
\section{Simple sensor intentions}
\label{sec:SSI}

The key idea behind simple sensor intentions can be summarized by the following principle: in the absence of an external reward signal, a sensible exploration strategy can be formed by learning policies that deliberately cause an effect on the observed sensor values by e.g. driving the sensor responses to their extremas or by controling the sensor readings at desirable set-points.

Clearly, policies that achieve the above can learn to cover large parts of the observable state-space, even without external rewards, and should thus be useful for finding `interesting' regions in the state-space. This idea is distinct from, but reminiscent of, existing work on intrinsic motivation for reinforcement learning \citep{gregor2016variational} which often defines some form of curiosity (or coverage) signal that is added to the external reward during RL. Our goal here is to learn separate exploration policies from general auxiliary tasks, that can collect good data for the main learning task at hand. 

As a motivating example, an analogy to this idea can be found by considering the exploration process of an infant: in the absence of `knowing what the world is about', a baby will often move its body in a seemingly `pointless', but goal directed, way until it detects a new stimulus in its sensors, e.g. a new sense of touch at the fingertips or a detected movement in a toy.

Simple sensor intentions (SSIs) propose a generic way to implement this principle in the multi-task agent framework SAC-X. SSIs are a set of auxiliary tasks, defined by standardized rewards over a set of scalar sensor responses. While many sensors in robotics naturally fit into this scheme (like e.g. a binary touch sensor or a sensor for a joint position), other sensors like raw camera images may need some transformation first in order to provide a scalar sensor signal that can be used to define a reasonable simple sensor intention.

In general, SSIs are derived from raw sensor observations in two steps:

\medskip
\noindent\emph{First step:} In the first step we map the available observations to scalar sensor responses that we want to control. Each observation $o \in s$ is a vector of sensor values coming from different sensory sources like e.g. proprioceptive sensors, haptic sensors or raw images.

We define a scalar (virtual) sensor response by mapping an observation $o$ to a scalar value $z$, i.e. 
$$
z = f(o), \text{where } o \in s,
$$
and where $f$ is a simple transformation of the observation. For scalar sensors, $f$ can be the identity function, while other sensors might require some pre-processing (like e.g. raw camera images, for which we describe a simple transformation in more detail in section \ref{sec:I2SSI}).
In addition -- as a consequence of choosing the transformation -- for each derived scalar sensor value $z$ we can also assume to know the maximum and minimum attainable value $[z_\text{min}, z_\text{max}]$.

\medskip
\noindent\emph{Second step:} In the second step we calculate a reward following one of two different schemes (described in detail below):
\begin{enumerate}
    \item[a)] rewarding the agent for reaching a specific target response, or
    \item[b)] rewarding the agent for incurring a specific change in response.
\end{enumerate}
Importantly, these schemes do not require a detailed semantic understanding of the environment: changes in the environment may have an a-priori unknown effect on sensor responses.
Irregardless of this relationship between environment and sensor values we follow the hypothesis outlined earlier: a change in a sensor response indicates some change in the environment and by learning a policy that deliberately triggers this change (a simple sensor intention, SSI) we obtain a natural way of encouraging diverse exploration of the environment.

\medskip
\subsection{Target Response Reward} \label{sec:TRR}
Let $z_t \in \{z_t \in \mathbb{R} \, | \, z_{\textrm{min}} \leq z_t \leq z_{\textrm{max}} \}$ be a sensor response $z_t = f(o)$ for observation $o \in s_t$. We define the reward at time step $t$ for controlling the response $z$ towards a desired set point $\hat{z}$ as

\begin{equation*}
r^{\hat{z}}(s_t) \coloneqq 1 - \frac{\abs{z_t - \hat{z}}}{z_{\textrm{max}} - z_{\textrm{min}}},
\end{equation*}

\noindent where $\hat{z}$ is the set point to be reached. The set point could be chosen arbitrarily in the range $[z_{\textrm{min}}, z_{\textrm{max}}]$, a sensible choice that we employ in our experiments, is to use the minimum and maximum response values as set points to encourage coverage of the sensor response space. We denote the two corresponding rewards as \emph{`minimize z'} and \emph{`maximize z'}, respectively in our experiments.


\medskip
\subsection{Response Change Reward} \label{sec:DRR}
While set point rewards encourage controlling sensor values to a specific value, response change rewards encourage the policy to incur a signed change of the response. Let $\Delta_t^z \coloneqq z_t - z_{t-1}$ be the temporal difference between consecutive responses. We define the reward for \emph{changing} a sensor value $z$ as
\begin{equation*}
r^\Delta(s_t) \coloneqq \frac{\alpha \Delta_t^z}{z_{\textrm{max}} - z_{\textrm{min}}}.
\end{equation*}
where $\alpha \in \{1, -1\}$ serves to distinguish between \emph{increasing} and \emph{decreasing} sensor responses.
In both cases, undesired changes are penalized by our definition. This ensures that a successful policy moves the response consistently in the desired direction, instead of exploiting positive rewards by developing a cyclic behaviour \citep{randlov1998learning, ng1999policy}.

\medskip
\section{Transforming Images to Simple Sensor Intentions \label{sec:I2SSI}}

In the following, we describe the transformation we use to obtain scalar sensor responses from raw camera images.
Cameras are an especially versatile and rich type of sensor for observing an agent's surroundings and are thus particularly interesting for our purposes.
Cameras typically deliver two dimensional pixel intensity arrays as sensor values. There are numerous ways to map these pixel arrays to one dimensional sensor responses, which can then be used as part of the simple sensor reward schemes.

\begin{figure}
    \centering
    \includegraphics[width=.92\linewidth]{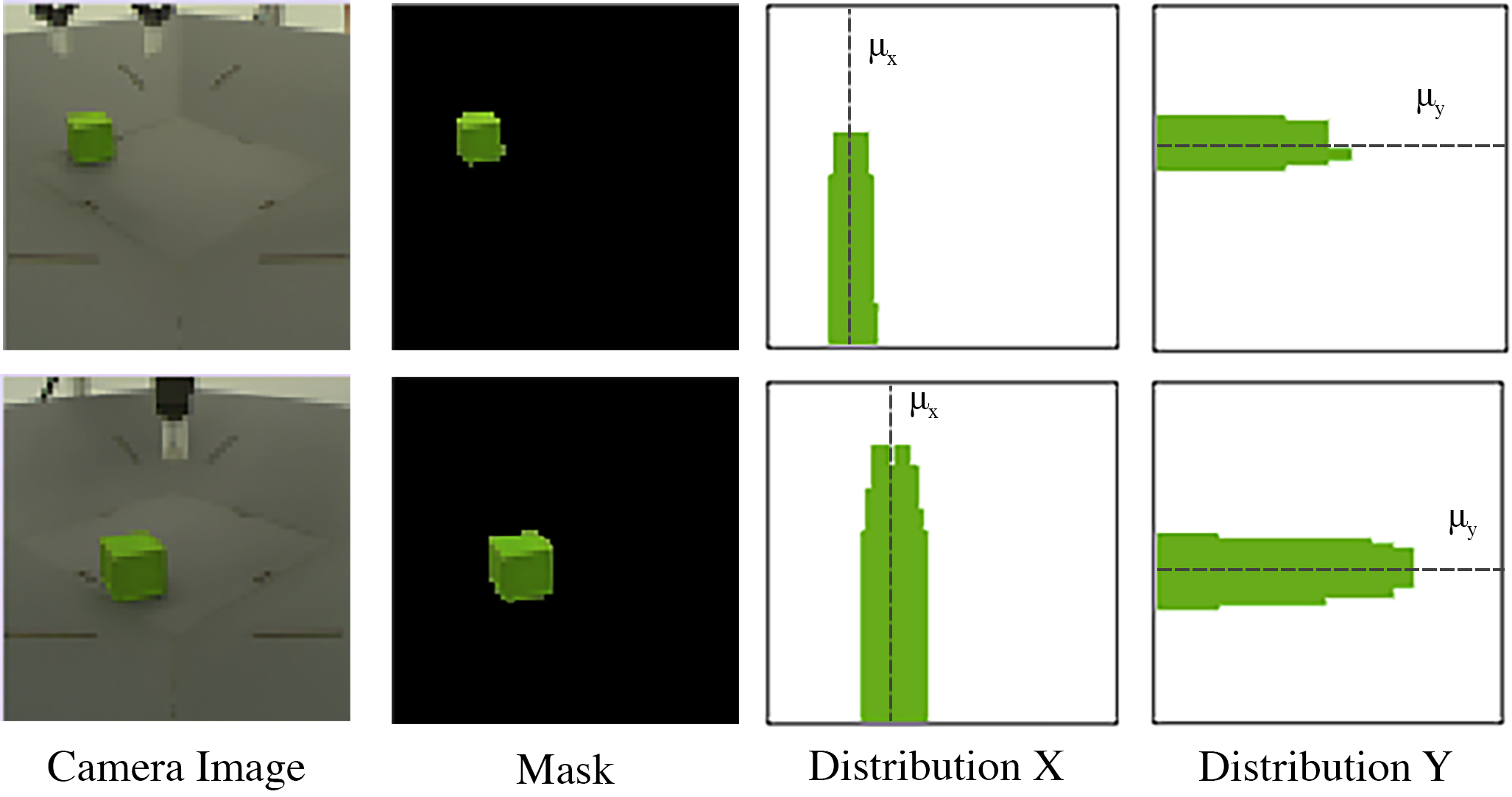}
    \caption{The transformation used for deriving one dimensional sensor responses from camera images. We compute a binary mask and it's spatial distribution along the image axes and select the resulting distribution's mean as a sensor response.}
    \label{fig:histogram}
\end{figure}


In principle, one could treat every pixel channel as an individual response, or calculate averages in regions of the image (similar to 'pixel control' in \citep{Jaderberg2017Unreal}), and subsequently define sensor rewards for each of the regions. However, our goal is to learn a sensor intention policy for e.g. maximizing / minimizing each sensor response, that can then be executed to collect data for a target task. In such a setting, having a smaller amount of efficient exploration policies is preferable (opposed to the large number of policies a reward based on single pixel values would mandate). We therefore propose to transform images into a small amount of sensor responses by aggregating statistics of an image's spatial color distribution. As illustrated in Figure \ref{fig:histogram}, we first threshold the image to retain only a given color (resulting in a binary mask) and then calculate the mean location of the mask along each of the image's axes, which we use as the sensor value. Formally we can define the two corresponding sensor values for each camera image as
\begin{equation*}
\begin{aligned}
z^{c_\text{range}}_x &= \frac{1}{W} \sum_{x=0}^W  x \max_y [\mathbf{1}_{c_\text{range}}(o_\text{image})[y, x]] \\
z^{c_\text{range}}_y &= \frac{1}{H} \sum_{x=0}^H  y \max_x [\mathbf{1}_{c_\text{range}}(o_\text{image})[y, x]],
\end{aligned}
\end{equation*}
where $H$ denotes the image height and $W$ the width, and $c_\text{range} = [c_\text{min}$, $c_\text{max}]$ correspond to color ranges that should be filtered.
Combined with the reward scheme above, these simple sensor responses can result in intentions that try to color given positions of an image in a given color. Perhaps surprisingly, we find such a simple reward to be sufficient for encouraging meaningful exploration in a range of tasks in the experiments.
We note that the reward formulation outlined in Section \ref{sec:SSI} mandates a sensor response to be available at each time step $t$. To avoid issues in case no pixel in the image matches the defined color range, we set any reward based on $z^\text{image}$ to zero at time $t=0$ and subsequently always fall back to the last known value for the reward if no pixel response matches. 

The choice of the `right' color range $c_\text{range}$, for a task of interest, is a design decision that needs to be made manually. In practice, we define a set of color ranges (and corresponding sensor values) from rough estimates of the color of objects of interest in the scene.
Alternatively, the color filters could be defined very broadly, which increases generality and transfer. For example, similar to a baby's preference for bright or vivid colors, one might use a color filter that matches a broad range of hues but only in a narrow saturation range. 
Furthermore, if the number of interesting color ranges is large (or one chooses them randomly) one can also define aggregate intentions, where sensor rewards are averaged over several sensor responses for various color channels, such that the resulting intention has the goal of changing an arbitrary color channel's mean instead of a specific one. In addition to manually defining color ranges, we as well conduct experiments with this aggregate approach.


\medskip
\section{Learning Simple Sensor Intentions for Active Exploration} \label{sec:sac}
In general, exploration policies based on the sensor rewards described in the previous section could be learned with any multi-task Reinforcement Learning algorithm or, alternatively, added as exploration bonus for an Reinforcement Learning algorithm that optimizes the expected reward for the target task $g$.

In this work we demonstate how simple sensor intentions can be used in the context of a multi-task RL algorithm, to facilitate exploration. We make use of ideas from the recent literature on data-efficient multi-task RL with auxiliary tasks (in our case defined via the simple sensory rewards). Concretely, we follow the setup from Scheduled Auxiliary Control (SAC-X) \citep{Riedmiller2018_Learning} and define the following policy optimization problem over $\mathcal{K}$ tasks:

\begin{equation*}
    \arg \max_{\pi} \mathbb{E}_{s \in \mathcal{B}} \Big[ \sum_{k=1}^K \mathbb{E}_{a \sim \pi_k(\cdot | s)} [Q^k_\phi(s, a)] \Big],
\end{equation*}
where $\pi_k(a | s)$ is a task-conditioned policy and $Q_\phi(s, a, k)$ is a task-conditional Q-function (with parameters $\phi)$; that is learned alongside the policy by minimizing the squared temporal difference error:
\begin{equation*}
    \min_{\phi} \mathop{\mathbb{E}}_{(s, a, s') \in \mathcal{B}} \Big[ \sum_{k=1}^K \big(r_k(s, a) + \gamma \mathbb{E}_{\pi_k} [Q^k_{\hat{\phi}}(s', a')] - Q^k_\phi(s, a) \big)^2 \Big],
\end{equation*}
where
$\hat{\phi}$ are the periodically updated parameters of a target network. We refer to the appendix for a detail description of the neural networks used to represent $\pi_k(a | s)$ and $Q_\phi(s, a, k)$.

The set of tasks $\mathcal{K}$ is given by the reward functions for each of the SSIs that we want to learn, as well as the externally defined goal reward $r_g$. The transition distribution, for which the policy and Q-function are learned, is obtained from the replay buffer $\mathcal{B}$, which is filled by \emph{executing both the policy for the target task $\pi_g$ as well as all other available exploration SSIs $\pi_k \in \mathcal{K}, k \neq g$}. In SAC-X, each episode is divided into multiple sequences and a policy to execute is chosen for each of the sequences. The decision which policy to execute is either made at random (referred to as Scheduled Auxiliary Control with uniform sampling or SAC-U) or based on a learned scheduler that maximizes the likelihood of observing the sparse task reward $r_g$ (referred to as SAC-Q). More details on the Reinforcement Learning procedure can be found in the Appendix as well as in the original Scheduled Auxiliary Control publication \citep{Riedmiller2018_Learning}.
%
%
%

\medskip
\section{Experiments}

In the following, simple sensor intentions (SSIs) based on basic sensors (like e.g. touch)
and more complex sensors, like raw camera images, are applied to several
robotic experiments in simulation and on a real robot. We show, that by using the concept of SSIs,
several complex manipulation tasks can be solved: grasping and lifting an object, stacking
two objects and solving a ball-in-cup task end-to-end from raw pixels. In all experiments, we assume the final task reward to be given in form of a sparse (i.e. binary) external reward signal.

\medskip
\subsection{Experimental Setup}
In all following experiments we employ a Rethink Sawyer robotic arm, with either a Robotiq 2F-85 parallel gripper as end-effector or, in the case of the Ball-in-a-Cup task, with a custom made cup attachment. In the manipulation setups, the robot faces a 20cm x 20cm basket, containing a single colored block, or - in the case of the stack task - two differently colored blocks. The basket is equipped with three cameras, that are used as the only exteroceptive inputs in all manipulation experiments (Figure \ref{fig:experimental_setup}, left). For the Ball-in-a-cup task, the robotic arm is mounted on a stand as shown in Figure \ref{fig:experimental_setup} (right). The Ball-in-a-cup cell is equipped with two cameras positioned orthogonally -- and both facing the robot -- as well as with a Vicon Vero setup. The Vicon system is solely used for computing a sparse external ''catch`` reward.

In all experiments we use Scheduled Auxiliary Control (SAC-X) with a scheduler choosing a sequence of 3 intentions per episode. Since the environment is initialized such that the initial responses are uniformly distributed, the expected return of the `increase' and `decrease' rewards following the optimal policy $\pi^*_k$ for a respective sensor reward is
\begin{equation*}
V^{\pi^*_k}(s) \leq \frac{|z^k_{\textrm{max}} - z^k_{\textrm{min}}|}{2} \quad \forall \: s \in \mathcal{S}.
\end{equation*}

\noindent Accordingly, we scale the increase and decrease rewards by a constant factor $\frac{2\sigma}{|z_{\textrm{max}} - z_{\textrm{min}}|}$ in all experiments and choose $\sigma = 200$ (which corresponds to the number of steps an SSI is executed in an episode) to achieve comparable reward scales across the reward schemes.

In all simulated setups we use uniform scheduling (SAC-U), where intentions are chosen by uniform random sampling \cite{Riedmiller2018_Learning}.  Also, we use a multi-actor setup with 64 concurrent actors that collect data in parallel. This approach optimizes for wall-clock time and sacrifices data-efficiency, which we accepted, since these experiments were primarily intended to answer the general question whether SSIs allow us to learn sparse tasks. The question of data-efficiency is relevant, however, in the real robot experiments and we thus used a jointly learned scheduler to select between policies to execute (SAC-Q, \cite{Riedmiller2018_Learning}) for improved data-efficiency. All real world experiments were conducted using a single robot (corresponding to a single-actor setup in simulation).

In all cases, the agent is provided with an observation that comprises proprioceptive information - joint positions, velocities and torques - as well as a wrist sensor's force and torque readings in the manipulation setups. Additionally, the agent receives the camera images as exteroceptive inputs. Thus, all tasks have to be learned from proprioceptive sensors and raw pixel inputs. The action space in the manipulation setups is five dimensional and continuous and consists of the three Cartesian translational velocities, the angular velocity of the wrist around the vertical axis and the speed of the gripper’s fingers. The action space in the Ball-in-a-Cup setup is four dimensional and continuous and consists of the raw target joint velocities \cite{schwab19simultaneously}. In all cases, the robot is controlled at 20 Hz.

\medskip
\subsection{Learning to grasp and lift in simulation}

\begin{figure}[t]
    \centering
    \includegraphics[width=.9\linewidth]{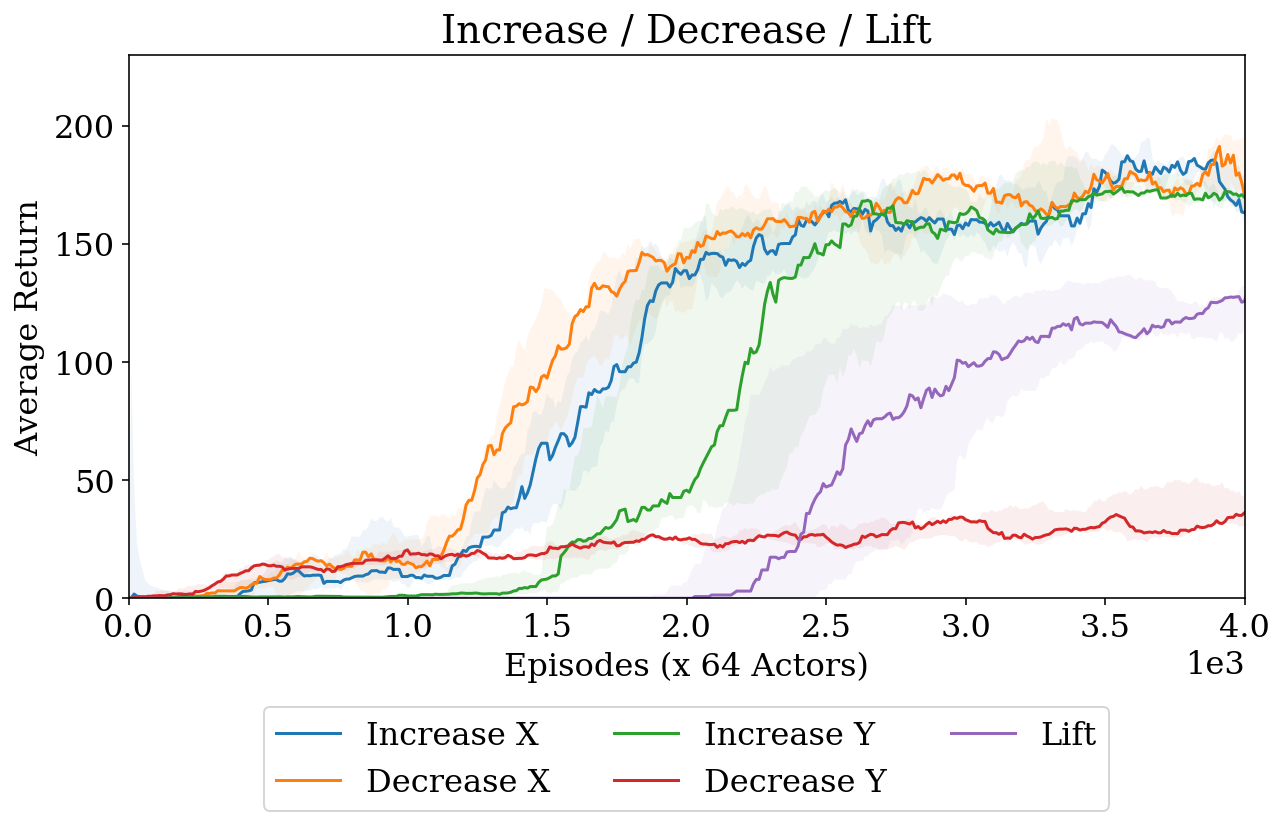}
    \caption{`Lift' learned from pixels in the simulated manipulation setup with the `increase` and `decrease` rewards used as auxiliary intentions.}
    \label{fig:inc_dec_lift_pixels_sim}
\end{figure}

Grasping and lifting an object with a robotic arm is a challenging task, in particular when learned from scratch and purely from pixels: The agent must learn to recognize the object, approach it, find the right position for grasping and eventually close the fingers and lift the object. Learning this from scratch, when only a sparse final reward is given, is very unlikely. 
We assume the external target reward for lift is given by a binary signal: the external reward is 1, if the touch sensor is triggered and the gripper is at least 15cm above the basket.

In a first experiment, we are using the 'delta response rewards' SSIs, that give reward for pushing the distribution of pixels of a selected color channel in a certain direction. In this experiment, we select the color channel to roughly match the color of the block. However, we will discuss below in section \ref{sec::ablations}, how we can get rid of this assumption, or make it more general, respectively. This selection of SSIs results in overall four auxiliary intentions, two for increasing the mean in x- or y-direction, and  two for decreasing the mean in x- or y-direction of the pixel image. The learning curves for the four auxiliaries and the final lift reward (violet line) are shown in Figure \ref{fig:inc_dec_lift_pixels_sim}. After about 500 episodes (times 64 actors), the agent sees some reward for moving the mean in various directions. This results in first, small interactions of the arm with the object. After about 1000 episodes, interactions get stronger, until after about 2000 episodes, the agent learns to lift the object deliberately. Final performance is reached after about 4000 episodes. We note that the reward curve for the `decrease y' intention is lower than the other curves because of the fact, that the block is typically at the bottom of the basket and therefore the mean of the color channel's distribution is typically already pretty low, so 'decrease y' is not able to earn as much reward as the other auxiliaries. This is an example of an intention that is potentially not very useful, but however does not prevent the agent of finally learning the goal task.

\begin{figure}[t]
    \centering
    \includegraphics[width=.9\linewidth]{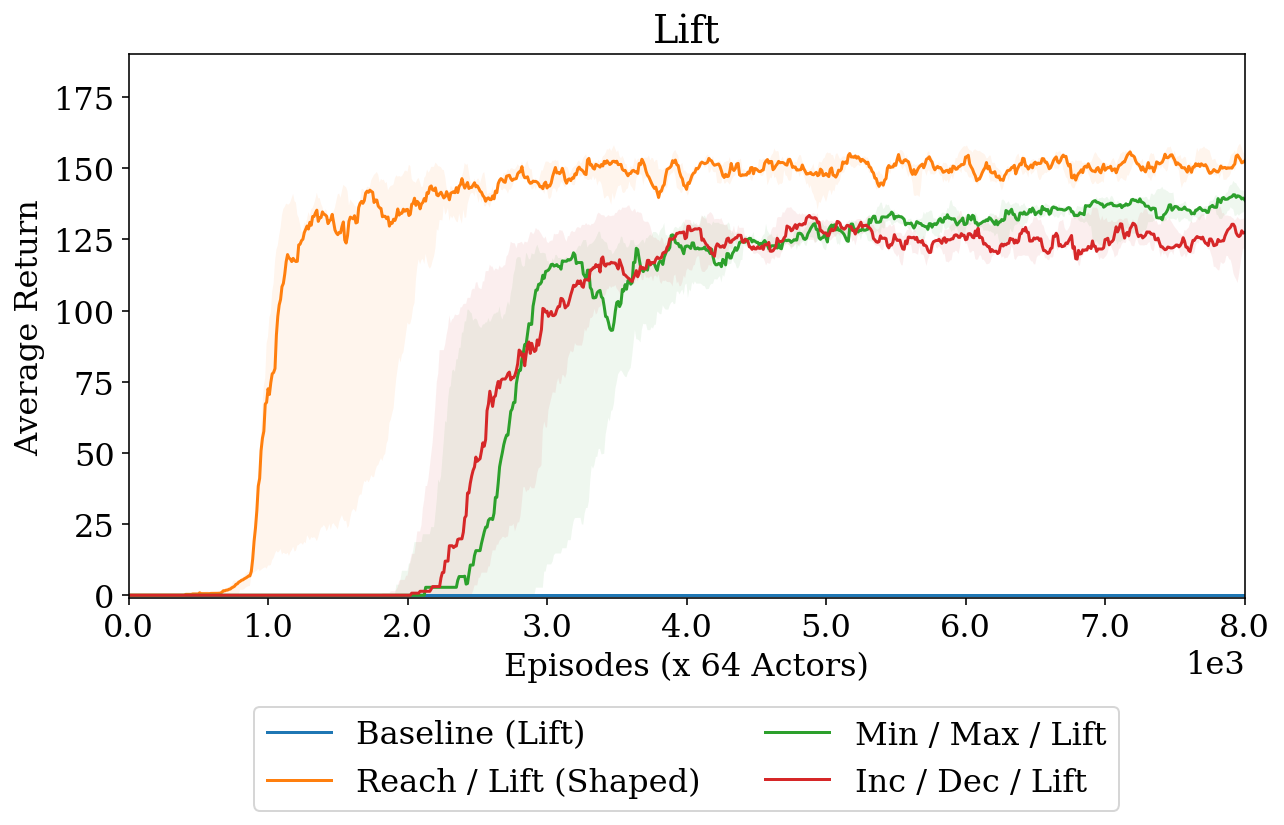}
    \caption{Comparison of the different reward schemes for learning `Lift` from pixels in the simulated manipulation setup.}
    \label{fig:lift_pixels_sim}
\end{figure}

If no auxiliary rewards are given, the agent does not learn to lift the object at all. This is shown by the flat blue learning curve in Figure 
 \ref{fig:lift_pixels_sim}. This Figure also shows the learning curve for the alternative `target response reward' SSI formulation, which tries to minimize/ maximize the value of the mean (green line) in comparison to the 'delta response reward' SSI formulation (red line). For the lift experiment, not much difference in the learning curves can be seen; both SSI formulations work successfully in this setting. For further reference, we also conducted a typical learning experiment using a dedicated shaping reward for reaching and grasping the block (orange line). As expected, the shaped approach learns faster, but at the costs of a considerable effort in the specification of the shaping reward and the required instrumentation for computing the position the object (which would require camera calibration, object detection, and pose estimation on a real system).
 
\subsection{Ablation studies \label{sec::ablations}}

We conducted several ablation studies and summarize our findings below:

\medskip
\subsubsection{Learning success does not depend on a particular camera pose for reward giving}

\begin{figure}
    \centering
    \centerline{
    \includegraphics[width=.25\linewidth]{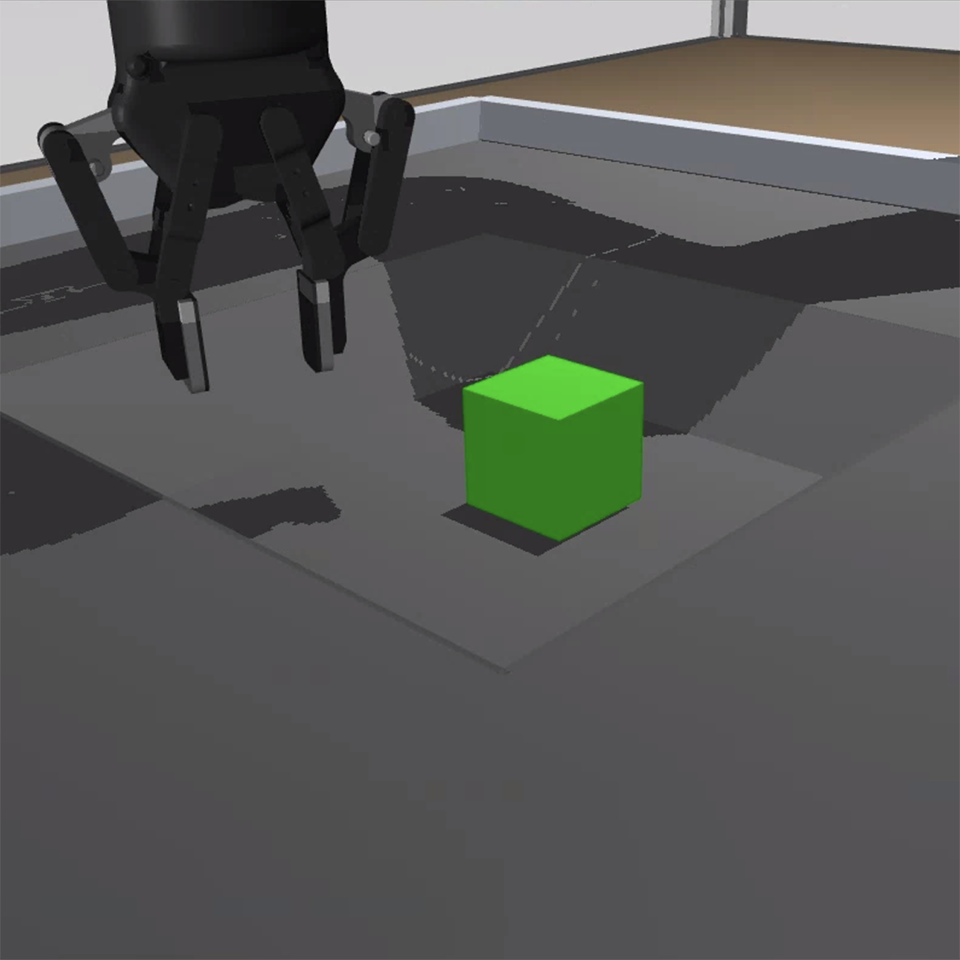} 
    \includegraphics[width=.25\linewidth]{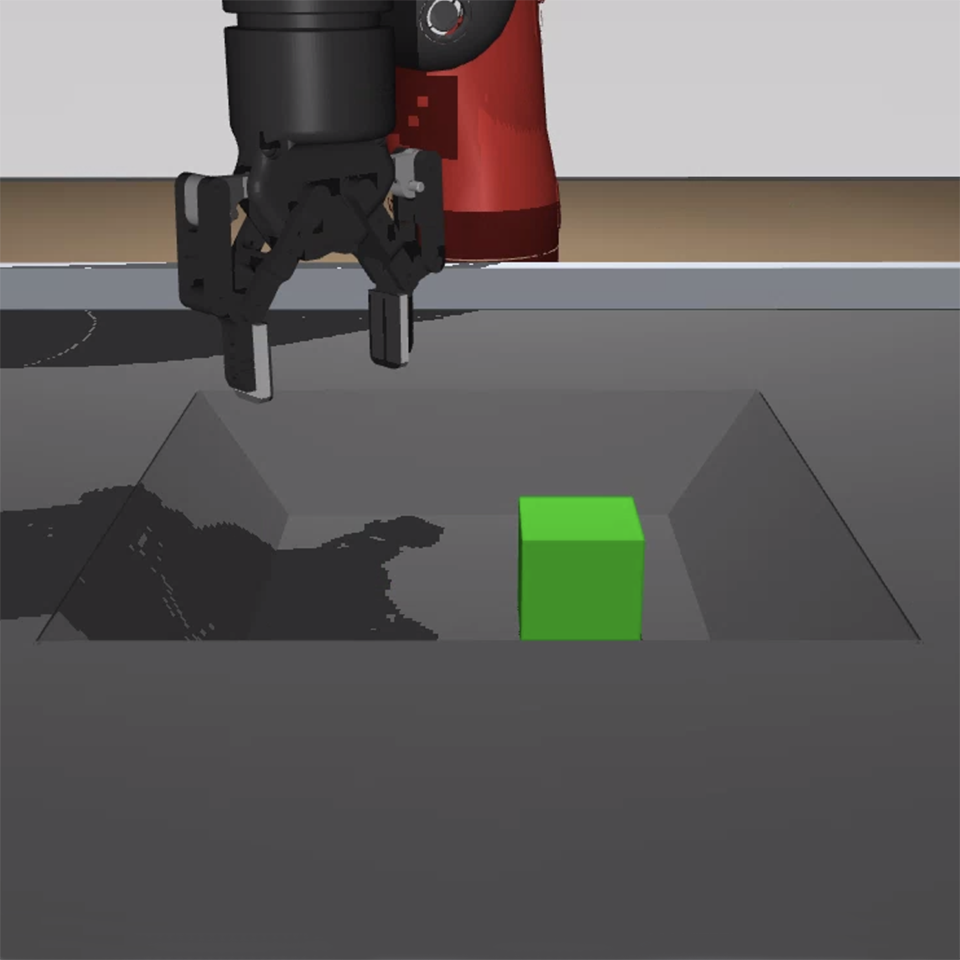} 
    \includegraphics[width=.25\linewidth]{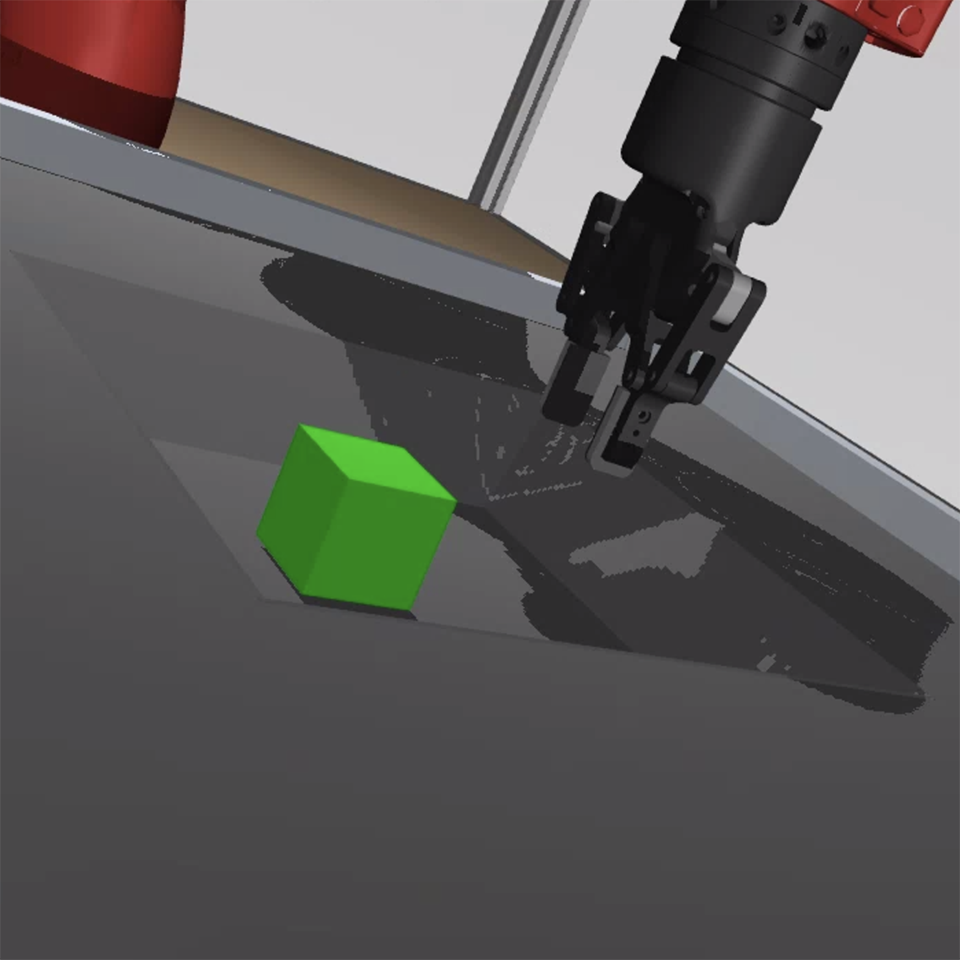} 
    }
    \caption{The three camera angles used for computing the SSIs for the experiments shown in Figure \ref{fig:lift_reward_cameras}.}
    \label{fig:camera_positions}
\end{figure}

\begin{figure}
    \centering
    \includegraphics[width=.9\linewidth]{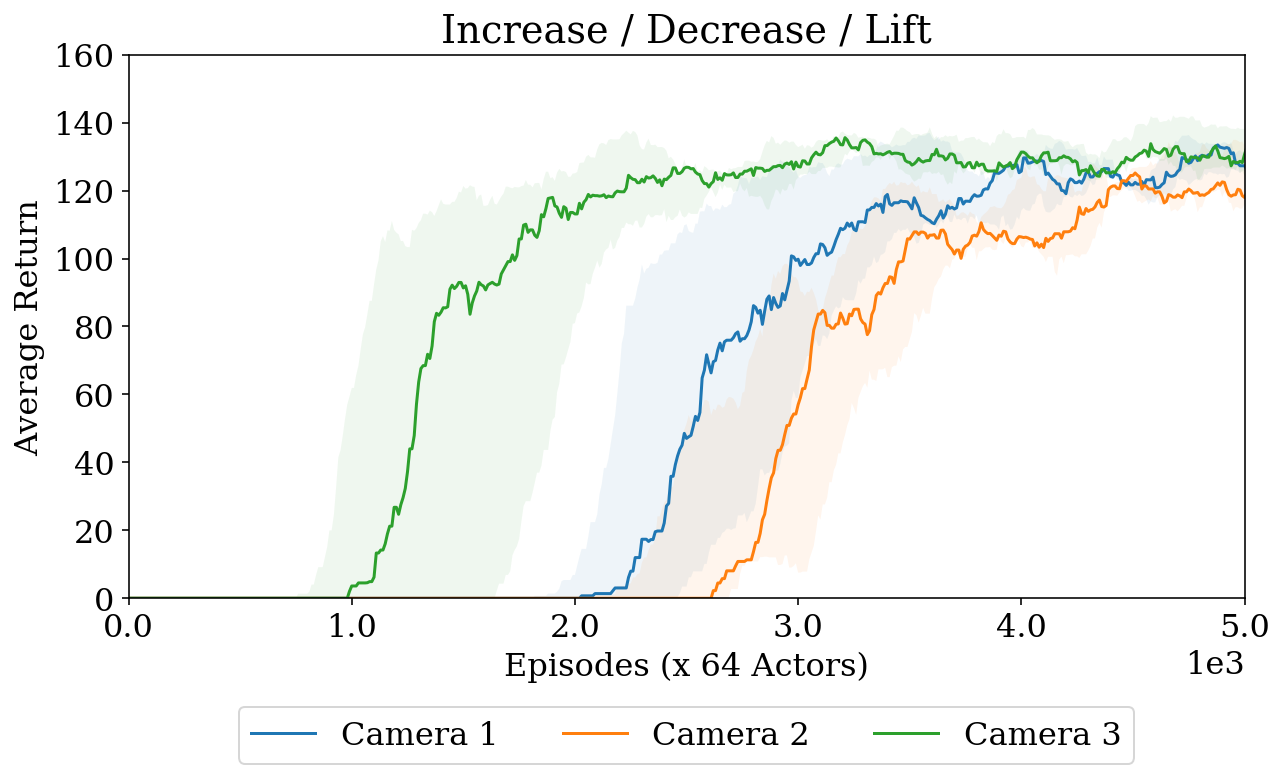}
    \caption{Comparison of different camera angles for computing SSIs when learning the `Lift' task. In all experiments the `increase' and `decrease' rewards are used as auxiliary intentions.}
    \label{fig:lift_reward_cameras}
\end{figure}

We investigated this by varying the perspective of the camera, that is used for reward computation, in various ways (see Figure \ref{fig:camera_positions}). As shown by the learning curves in Figure \ref{fig:lift_reward_cameras}, the concrete pose of the camera has an influence on learning speed, but in all cases, the SSIs were powerful enough to learn the final lifting task eventually.

\medskip
\subsubsection{The SSI color channel does not necessarily need to specify a single object}

\begin{figure}[b!]
    \centering
    \centerline{
    \includegraphics[width=.36\linewidth]{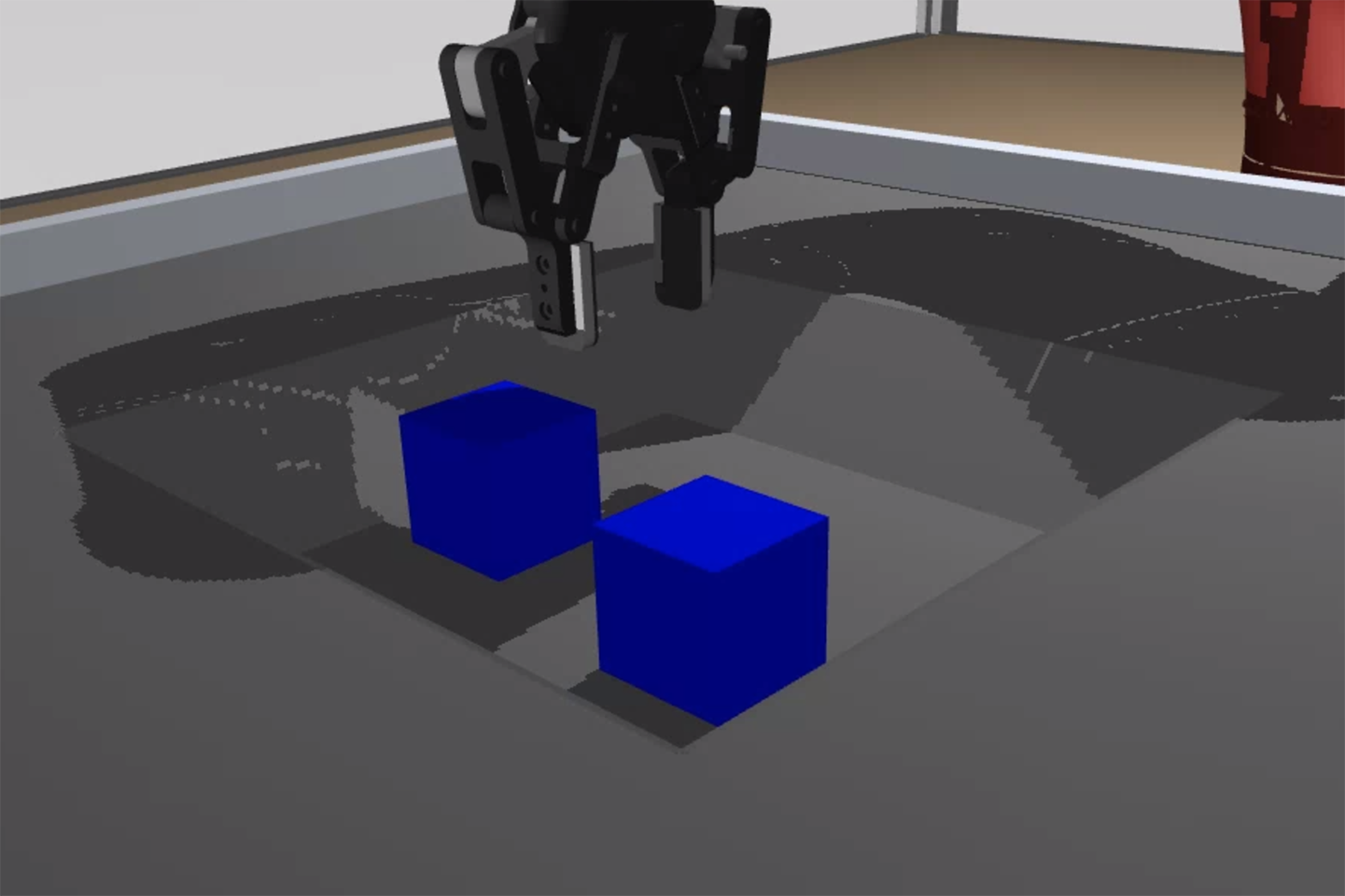} 
    \includegraphics[width=.36\linewidth]{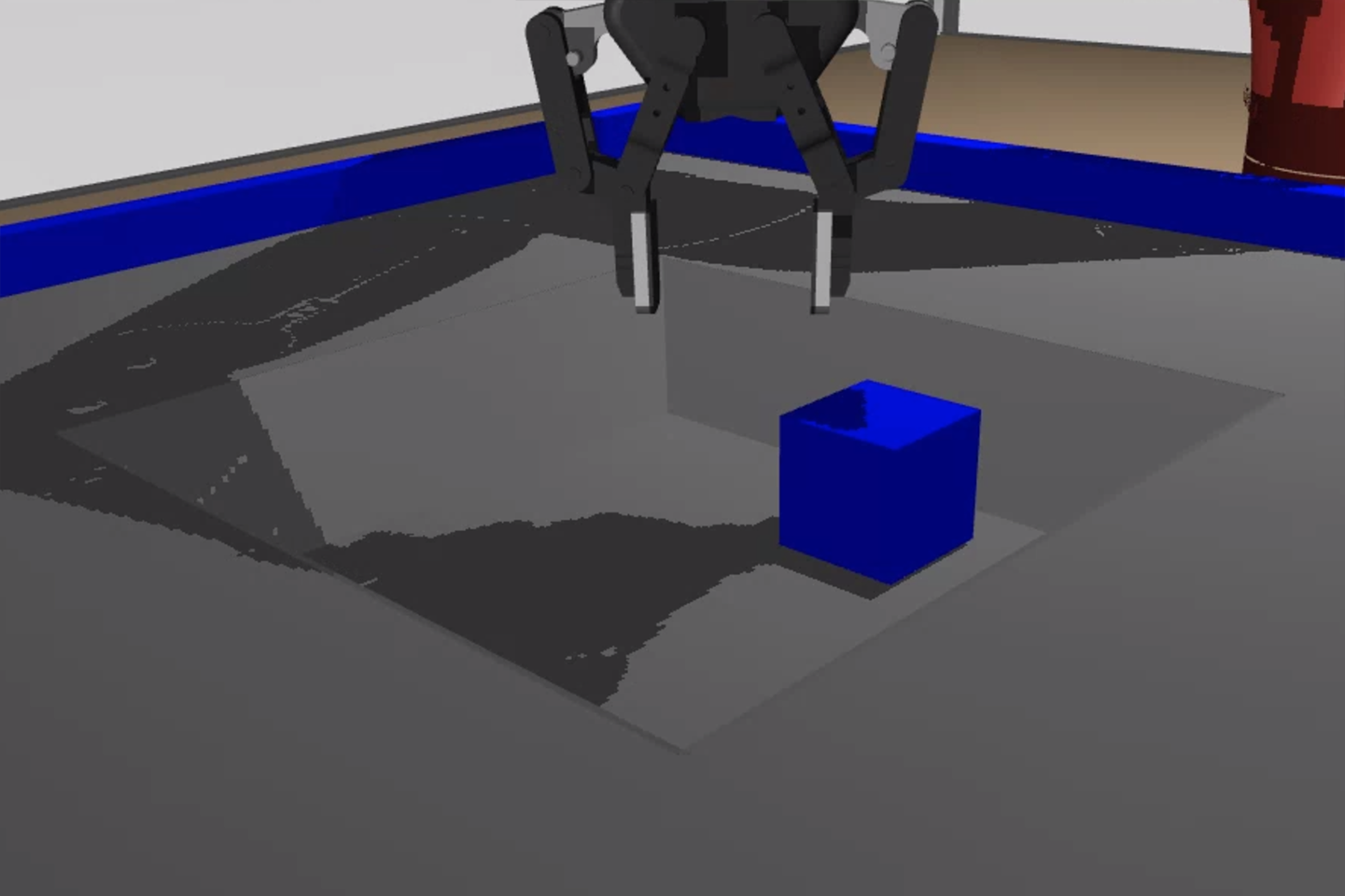}
    }
    \caption{The setups used for demonstrating robustness. On the left, both objects in the scene have the same color. On the right a non-moveable part of the basket has the same color as the object.}
    \label{fig:color_mask_ablations}
\end{figure}

We investigated this by two experiments: an experiment with two blocks of the same color (see figure \ref{fig:color_mask_ablations}, left) and an experiment, where part of the background had the same color as the block (see Figure \ref{fig:color_mask_ablations}, right). In both cases lift could be learned successfully using the standard pixel based SSIs. If we increase the 
proportion of the background pixels with the same color as the object, at some point the agent fails to learn to interact with the brick, but instead 'exploits' the reward scheme by trying to move the mean by hiding pixels with the arm and gripper. It is possible to filter the background from the scene; however, this is beyond the scope of this paper and therefore left for future investigations.


\medskip
\subsubsection{One can use a much more general selection of the color channel}

\begin{figure}[t]
    \centering
    \includegraphics[width=.36\linewidth]{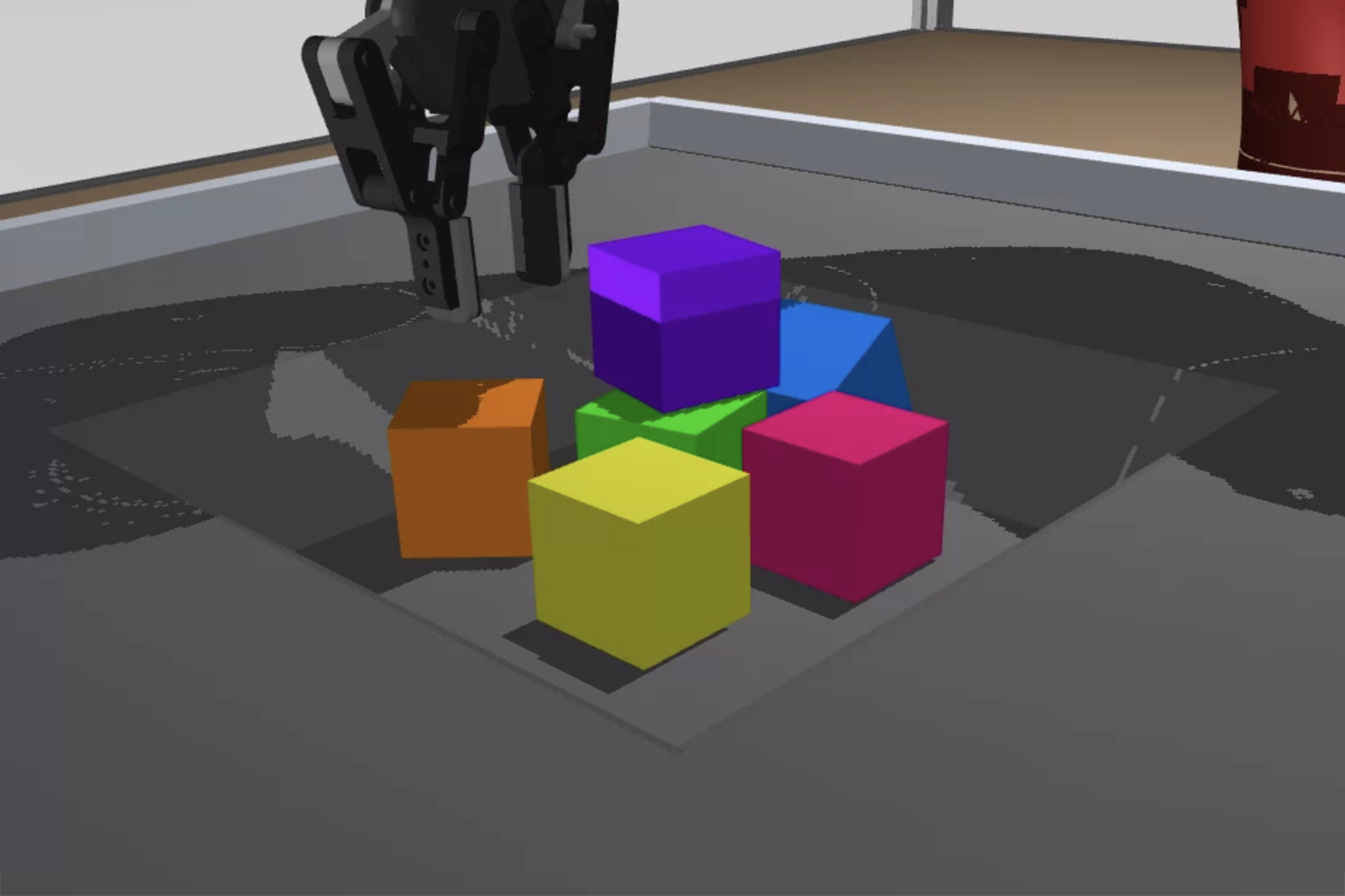}
    \includegraphics[width=.36\linewidth]{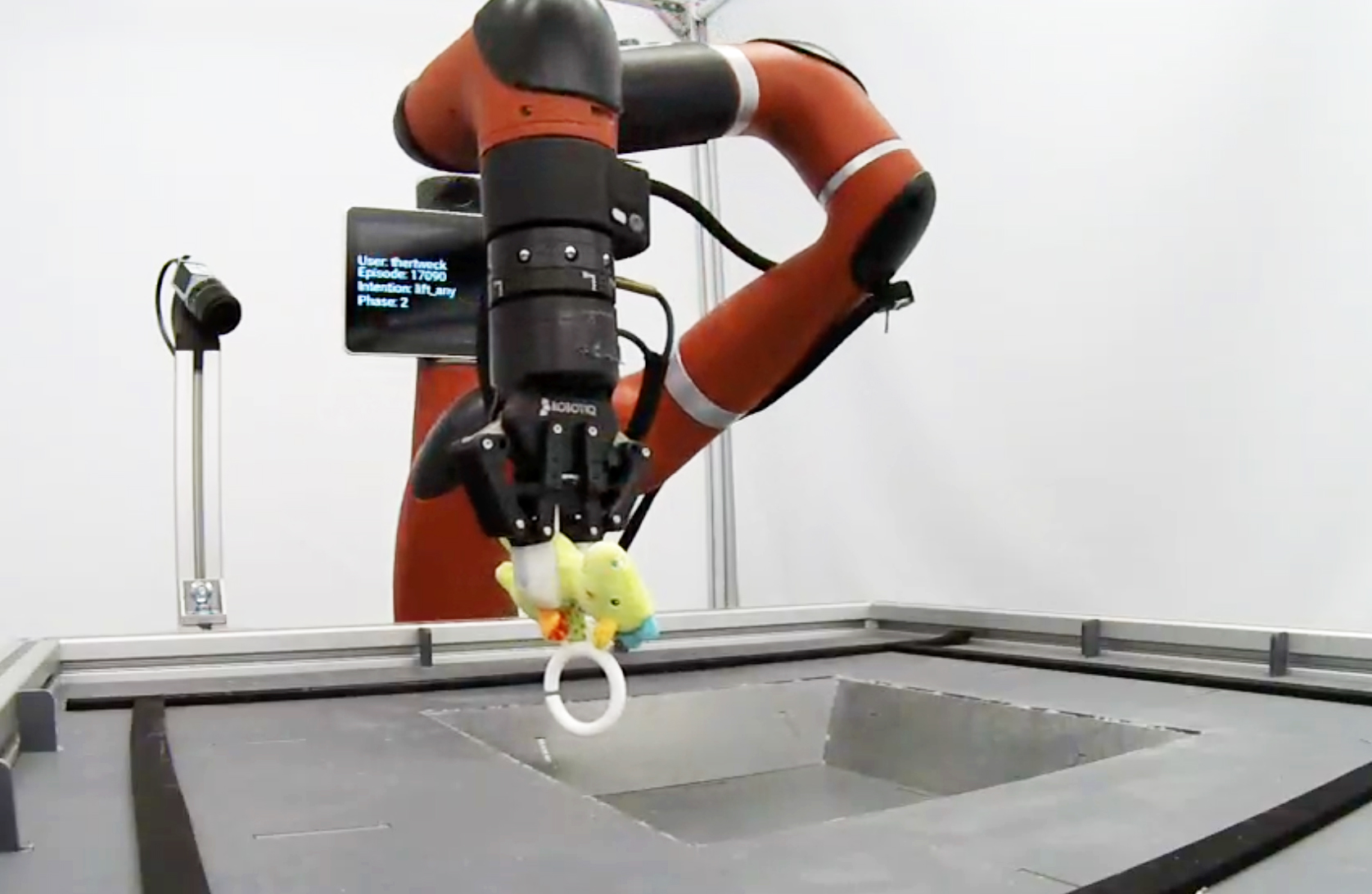}
    \caption{Left: A subset of the colored blocks used in the simulated `Lift Any' experiments. In each episode one of the available blocks is selected. Right: Evaluation of the `Lift Any' experiment showing that the robot is able to grasp and lift a non-rigid, multi-colored baby toy.}
    \label{fig:multi_color_bricks}
\end{figure}

\begin{figure}[t]
    \centering
    \includegraphics[width=.9\linewidth]{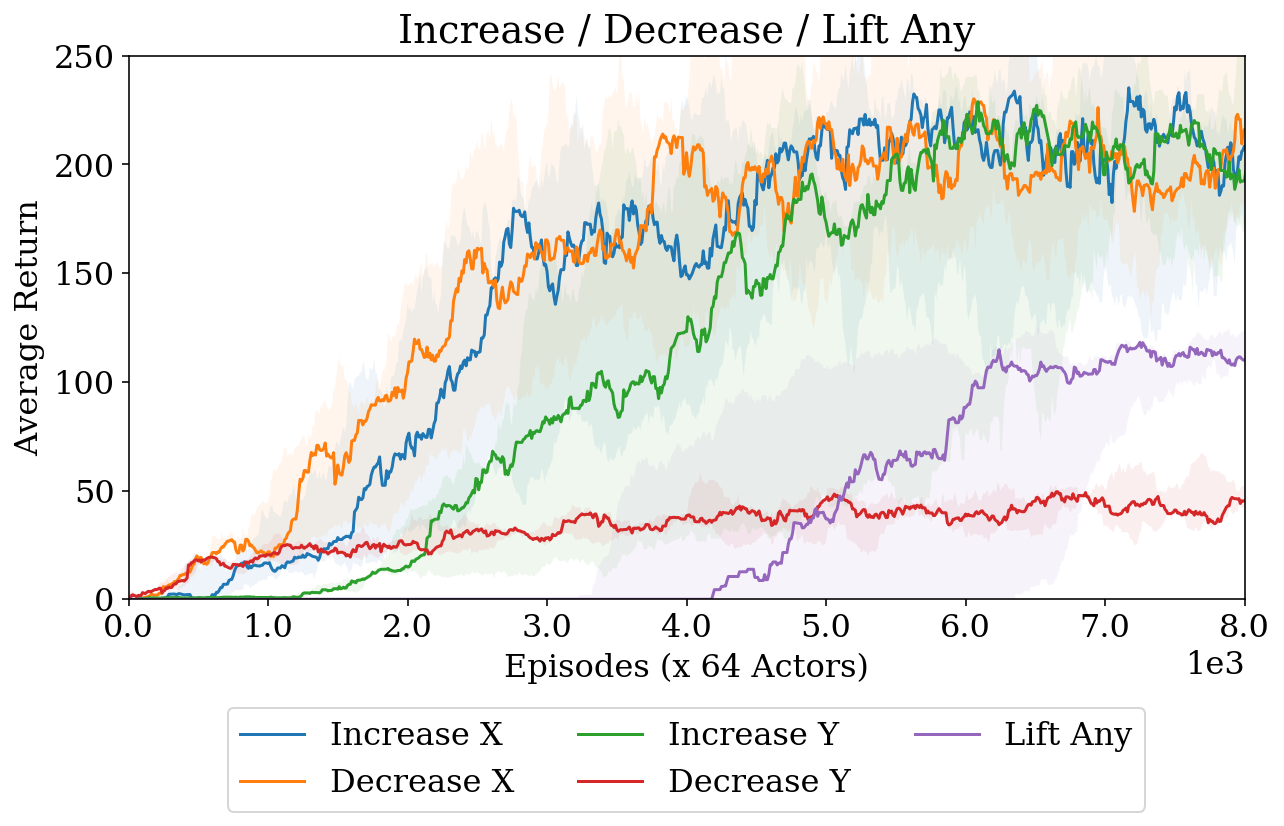}
    \caption{`Lift Any' learned from pixels in the simulated manipulation setup with the `increase' and `decrease` rewards used as auxiliary intentions. Here, the `increase' and `decrease' rewards are aggregated over color channels and the block's color is changed every episode.}
    \label{fig:lift_any_pixels_sim}
\end{figure}

To demonstrate this, we used the `aggregate' SSI formulation suggested above: We compute rewards (here: delta rewards) for a potentially large set of different color channels and add those up in one so-called `aggregate' SSI. This will clearly solve the single block lift task as described above, as long as at least one of the color channels matches the color of the block. However, this works also in a setting, where blocks with different colors are used throughout the experiment. Figure \ref{fig:lift_any_pixels_sim} shows the learning curve for the `Lift Any' task (red line), where we randomly changed the color of the block in every episode (see Figure \ref{fig:multi_color_bricks} showing a subset of the colored blocks used in the experiment).
A similar experiment was also conducted on the real robot. The results for the real world experiment is shown in Figure \ref{fig:lift_any_pixels_real}.

\medskip
\subsubsection{The SSI method is not restricted to pixels, but works with a general set of (robot) sensors}

In particular, we conducted experiments in a setup with two blocks in the scene, where we applied SSIs to basic sensors, like the touch sensor and the joint angles, but also used the SSIs described before on camera images, resulting in a total of 22 auxiliary intentions (minimize / maximize touch, minimize / maximize joint angles of arm and gripper joints, minimize / maximize the mean of the color distribution along the x and y axes). To deal efficiently with the extended set of intentions, the agent used a Q-based scheduler (SAC-Q), which successfully learned the `Lift' task.

\medskip
\subsubsection{It is mandatory to have a penalty term for moving the sensor response in the opposite direction}

In learning experiments, where the agent only received a positive reward, if the response is moved in the intended direction, the agent quickly learns to cleverly exploit the reward, e.g. by moving the gripper back and forth in front of camera, hiding and revealing the block and thereby collecting reward for moving the response (see \ref{sec:TRR} and \ref{sec:DRR}).

\medskip
\subsection{Learning to stack}

Learning to stack a block on another block poses additional challenges compared to the `grasp-and-lift' scenario described above: The scene is more complex since there are two objects now, reward is given only if the object is placed above the target object, and the target object can move. The external task reward for `stack' is 1, if and only if the block 1 is properly placed on block 2 and the gripper is open.

\begin{figure}
    \centering
    \includegraphics[width=.9\linewidth]{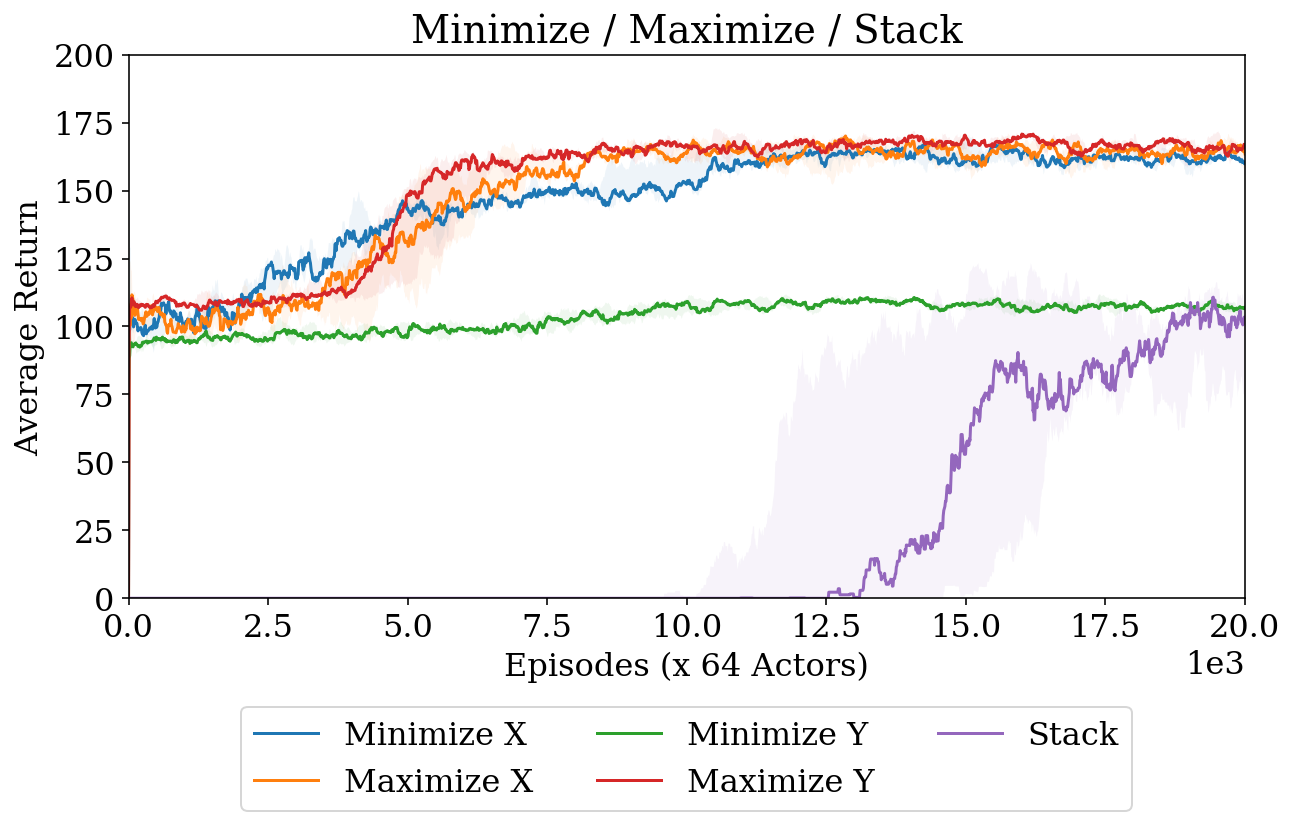}
    \caption{`Stack' learned from pixels in the simulated manipulation setup with the `minimize' and `maximize' rewards used as auxiliary intentions.}
    \label{fig:stack_pixels_sim}
\end{figure}

We use the minimize/ maximize SSI approach in combination with a SAC-U agent: the SSI auxiliary rewards are computed from raw pixels and the input observations for the agent comprises raw pixels and proprioceptive information only.
Figure \ref{fig:stack_pixels_sim} shows the learning curves for the auxiliaries and the final `Stack' reward. This is a much more difficult task to learn from pure pixels, but after about 20,000 episodes (times 64 actors) the agent has figured out how to reliably solve the task, purely from pixels. Not surprisingly, without SSI auxiliaries, the agent is not able to learn the task. The minimize / maximize auxiliary rewards are position based rewards - therefore the learning curves are offset by roughly the average reward of 100 which corresponds to the usual average position of the pixel mean.

Although the amount of data used to learn this task (20,000 episodes times 64 actors) is huge, it is still surprising, that learning such a complex task is possible from raw sensor information and an external task reward only. We assume this is possible, since the used simple sensor intentions encourage a rich playing within the scene, and the additional sequencing of these intentions as performed by SAC-X increases the probability of seeing also more complex configurations - like stacking in this case. From seeing external rewards for these occasional stacks, learning finally can take off.

\medskip
\subsection{Learning to grasp and lift on a real robot}

To investigate the behaviour of SSI based agents in a real world robotic setup, we apply the agent to learn to grasp and lift objects (figure \ref{fig:experimental_setup}, left). As in the simulation experiments, the agent's input is based on raw proprioceptive sensory information and the raw images of two cameras placed around the basket. Real world experiments add the additional challenge of noisiness of sensors and actuators, with which the agent has to cope. Also, the approach naturally has to work in a single actor setup, since there is only one robot available to collect the data.

For the real robot experiment, we use 6 SSIs based on the touch sensor as well as on the camera images: increase/ decrease of the color distribution's mean in x- and y-direction of raw camera images (4 intentions) plus minimize / maximize touch sensor value ('on' / 'off', 2 intentions). The task reward for `Lift' is given sparsely, if and only if the touch sensor is activated while the gripper is 15 cm above the table. To make the SSI reward signal from pixels more robust, we first computed the SSI values from each camera image and then aggregated the rewards to get one single accumulated reward. This means that an auxiliary task receives the maximum reward only if it achieves to move the color distribution's mean in both cameras in the desired direction.

\begin{figure}
    \centering
    \includegraphics[width=.9\linewidth]{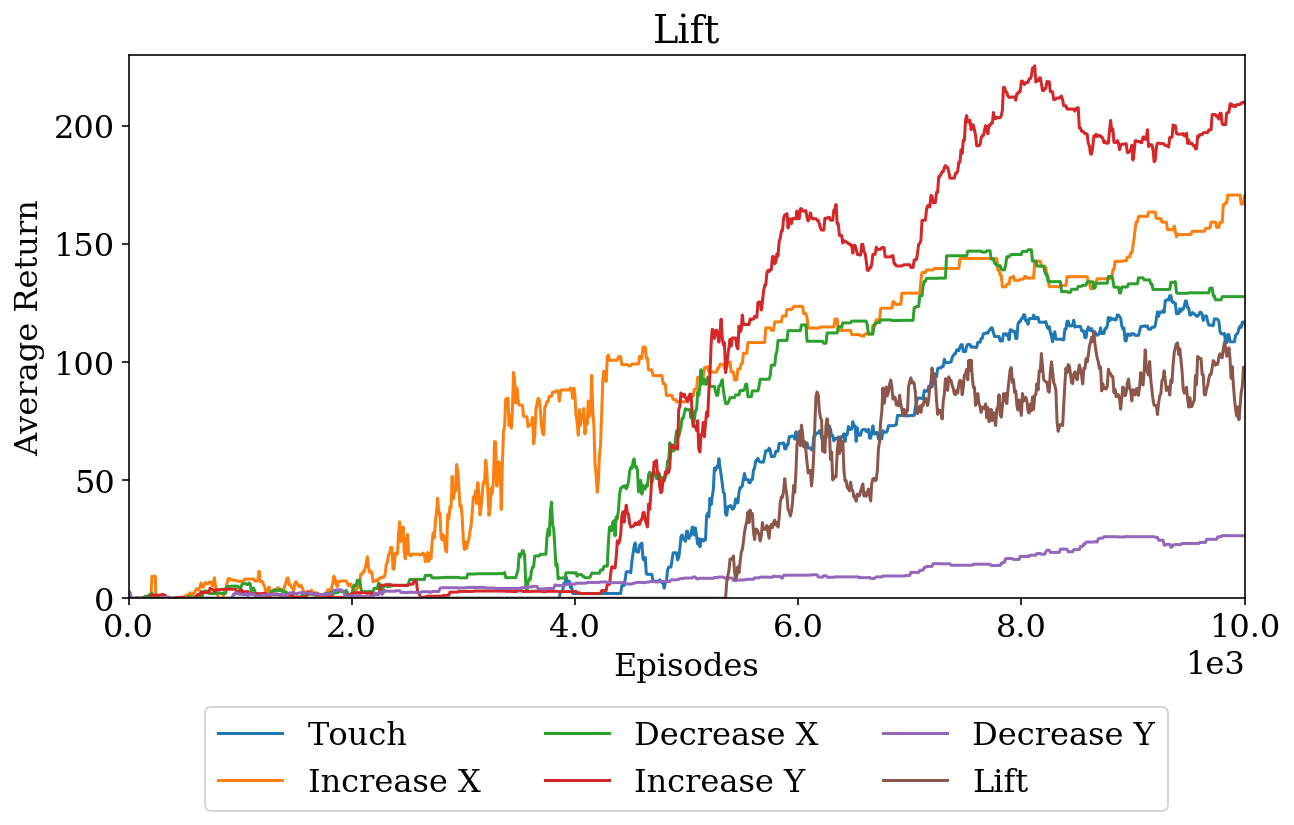}
    \caption{`Lift' learned from pixels on the real robot with the `increase' and `decrease' rewards used as auxiliary intentions. In this setup, the `increase' and `decrease' rewards are aggregated over two perpendicular camera angles.}
    \label{fig:lift_pixels_real}
\end{figure}

The learning agent is again based on Scheduled Auxiliary Control and we apply a Q-table based scheduler (SAC-Q) \cite{Riedmiller2018_Learning} to make learning as data-efficient as possible. We also add a special exploration scheme based on `multi-step' actions \cite{SR02, SR03:NCAF}: when selecting an action, we additionally determine how often to repeat its application, by drawing a number of action repeats uniformly in the range $[1, 20]$. Also, for safety reasons, external forces are measured at the wrist sensor and the episode is terminated if a threshold of 20N on any of the three principle axes is exceeded for more than three time steps in a row. If this happens, the episode is terminated with zero reward, enforcing the agent to avoid these failure situations in the future.

We find that the agent successfully learns to lift the block as illustrated in Figure \ref{fig:lift_pixels_real}. After an initial phase of playful, pushing-style interaction, the agent discovers a policy for moving the block to the sides of the basket and after exploring the first successful touch rewards, it quickly learns to grasp and lift the block. After about 9000 episodes the agent shows a reliable lifting behaviour. This corresponds to roughly 6 days of training on the real robot.

\begin{figure}
    \centering
    \includegraphics[width=.9\linewidth]{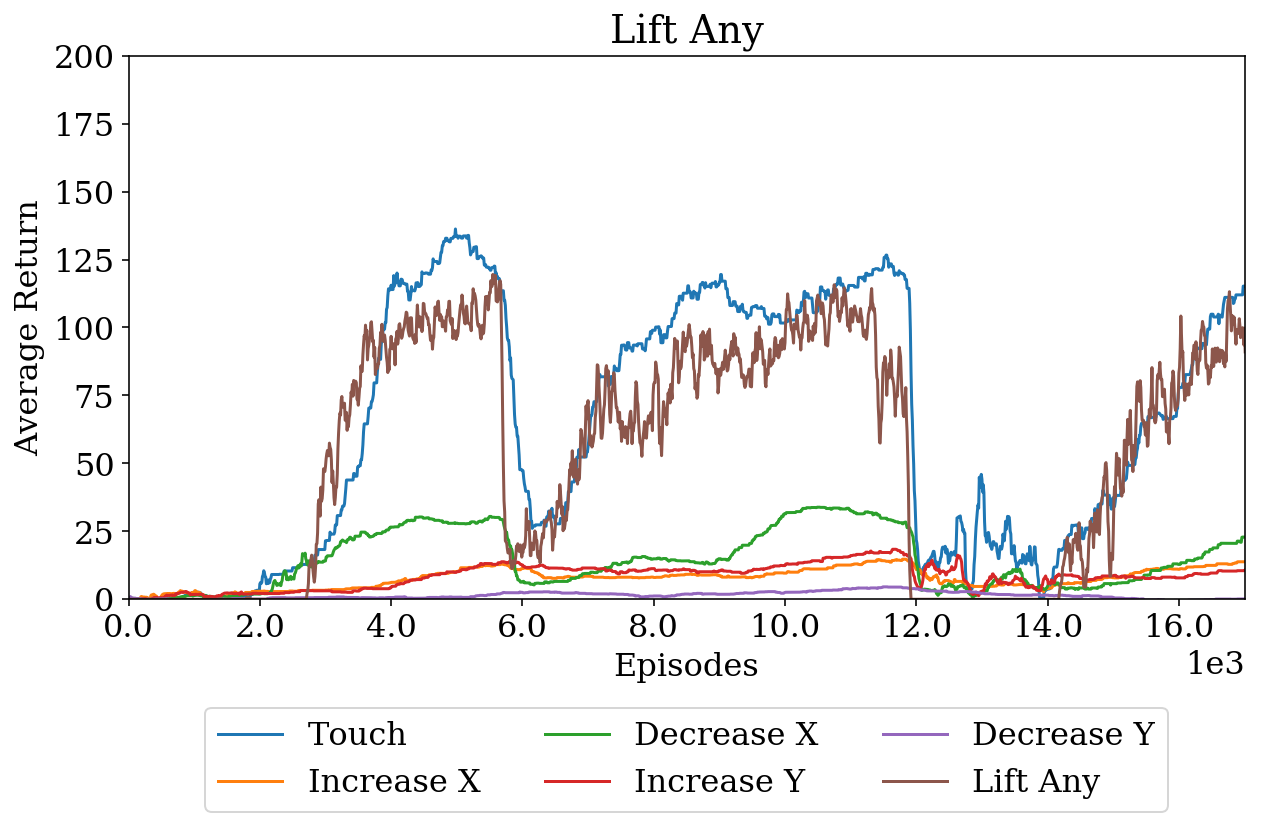}
    \caption{`Lift Any' learned from pixels on a real robot with the `increase' and `decrease' rewards used as auxiliary intentions. The block is replaced at various points throughout the experiment.}
    \label{fig:lift_any_pixels_real}
\end{figure}

In a further experiment, we replaced the above SSIs for a single color channel with SSIs, that aggregate rewards over multiple color channels, allowing to learn with objects of any color. We tested this by starting to learn with a single object, until the robot started to lift, and then replacing the object by another object of different color and/ or of different shape. The learning curve is shown in figure \ref{fig:lift_any_pixels_real}. The drops in the learning curve indicate, that if the object is replaced, the agent does not know how to lift yet. After some time of adaptation, it starts to manage to lift again. Continuing this, we saw the robot being able to learn to lift a wide variety of different objects, all from raw sensor information.

\medskip
\subsection{Ball in a Cup}

An important aspect of simple sensor intentions is, that SSIs constitute a basic concept and show some generality of being helpful for different external target tasks. To illustrate how the same set of SSIs can be employed to master a completely different control problem, we show results on the dynamic Ball-in-a-Cup task \cite{schwab19simultaneously}: the task is to swing up a ball attached to a robot arm and to catch it with a cup mounted as the end effector of the arm. The agent only receives a binary positive reward, if the ball is caught in the cup (see figure \ref{fig:experimental_setup}, right).

Dynamic tasks in general exhibit additional difficulties compared to static tasks, e.g. the importance of timing and the difficulty of reaching (and staying in) possibly unstable regimes of the robot's configuration-space. As a result, learning to catch the ball purely from pixels is out-of-reach for learning setups, that only employ the sparse catch reward.

\begin{figure}
    \centering
    \includegraphics[width=.9\linewidth]{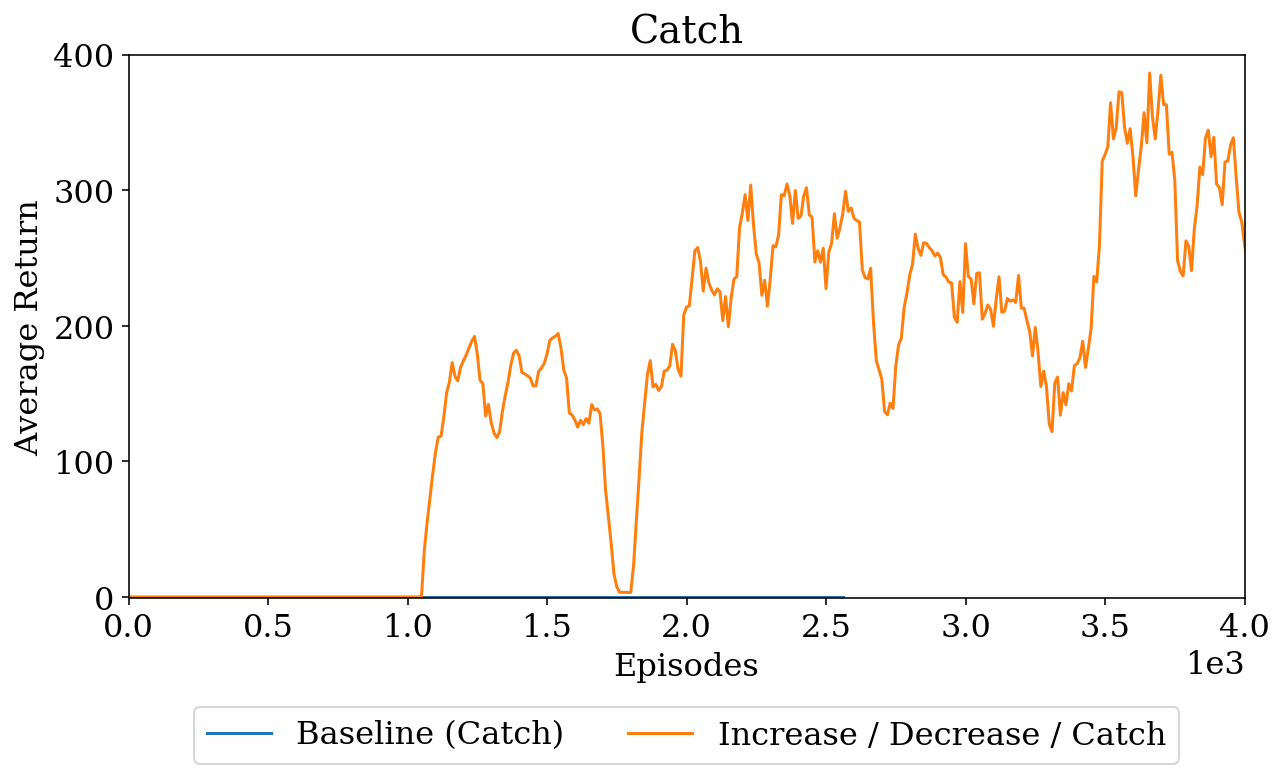}
    \caption{`Catch' learned from pixels on a real robot.}
    \label{fig:catch_real}
\end{figure}

To show the versatility of simple sensor intentions, we choose the standard increase / decrease SSIs, resulting in 4 auxiliary tasks plus
the binary task reward for catch. The learning agent used is a 
SAC-Q agent.
As shown in Figure \ref{fig:catch_real}, the agent is able to learn this dynamic task purely from the standard observation, pixels and proprioceptive inputs. In order to cope with the dynamic nature of the task, 2 consecutive pixel frames were stacked for the controller's input.

The simple sensor intentions encourage to move the colored pixels in the image, which results in learning to deliberately move the ball. The sequential combination of different auxiliary tasks as enforced by the SAC-X agent, lead to a broad exploration of different movements of ball and robot. Eventually, first catches are observed, and once this happens, the agent quickly learns the `Catch' task. For the roughly 4000 episodes needed, the learning took about 3 days in real time.

\section{Related Work}

Transfer via additional tasks has a long-standing history in reinforcement learning to address exploration challenges and accelerate learning \citep{torrey2010transfer,pan2010survey}.
We are able to conceptually distinguish into two main categories with auxiliary tasks \citep{taylor2009transfer,Jaderberg2017Unreal}, which are used to accelerate training on a final task, and multitask learning \citep{caruana1997multitask} with focus on the performance of all involved tasks.

Early work on multitask learning \citep{Dayan93} introduced the prediction of expected sums of future values for multiple tasks of interest. 
Many successive approaches have extended modelling of multiple task rewards and investigated further types of transfer across tasks \citep{Sutton2011horde, Schaul15, Barreto2017}.
Similarly, work on the options framework investigates directions for decomposition of task solutions \citep{Dietterich98, Bacon17,daniel2016hierarchical, wulfmeier2019regularized}.

Auxiliary tasks have been investigated as manually chosen to help in specific domains \citep{Riedmiller2018_Learning, Dosovitskiy2017, Jaderberg2017Unreal,Mirowski16,cabi2017} and as based on agent behaviour \citep{andrychowicz2017hindsight}. As manual task design presents significant burden for human operators, these methods demonstrate that even simpler sets of tasks can demonstrate considerable benefits. In comparison to methods using auxiliary tasks mostly for representation shaping by sharing a subset of network parameters across tasks  \citep{Jaderberg2017Unreal, Mirowski16}, SSI shares data between tasks which directly uses additional tasks for exploration.

For both directions, success considerably relies on the source of tasks and correlated reward functions, which this work focuses on. 
Multiple agent behaviour based sources for tasks and additional objectives have been considered in prior work including diversity of behaviours
\citep{sharma2019dynamics,grimm2019disentangled, eysenbach2018diversity},
intrinsic motivation \citep{Chentanez04,Singh09,blaes2019control, Ngo2012, Kulkarni2016}, 
and empowerment of the agent \citep{klyubin2005all, mohamed2015variational,houthooft2016vime}.
Additionally, representation learning and the automatic identification of independently controllable features have provided another perspective on identifying tasks for transfer and improving exploration \citep{grimm2019disentangled, bengio2017independently, blaes2019control}.
Recent work on diversity has in particular demonstrated the importance of the space used for skill discovery \citep{sharma2019dynamics}. SSI provides a valuable perspective on determining valuable task spaces with limited human effort.

Given large sets of tasks, the higher-level problem of choosing which task to learn creates a novel challenge similar to exploration within a single task. Often random sampling over all possible tasks provides a strong baseline \citep{graves2017automated}. However, to accelerate learning different perspectives build on 
curriculum \citep{bengio2009curriculum, heess2017emergence, Oudeyer17},
iterative task generation \citep{Schmidhuber2013PowerPlayTA,wang2019paired}. In this work, we rely on task scheduling similar to \citet{Riedmiller2018_Learning} in order to optimize the use of training time.

\section{Conclusion and future work}

Learning to change sensor responses deliberately is a promising exploration principle in settings, where it is difficult or impossible to experience an external task reward purely by chance. We introduce the concept of simple sensor intentions (SSIs) that implements the above principle in a generic way within the SAC-X framework. While the general concept of SSIs applies to any robotic sensor, the application to more complex sensors, like camera images is not straight forward. We provide one concrete way to implement the SSI idea for camera images, which first need to be mapped to scalar values to fit into the proposed reward scheme. We argue that our approach requires less prior knowledge than the broadly used shaping reward formulation, that typically rely on task insight for their definition and state estimation for their computation.

In several case studies we demonstrated the successful application to various robotic domains, both in simulation and on real robots. The SSIs we experimented with were mostly based on pixels, but also touch and joint angle based SSIs were used. The definition of the SSIs was straight-forward with no or minor adaptation between domains.

Future work will concentrate on the extension of this concept in various directions, e.g. improving the scheduling of intentions to deal with a large number of auxiliary tasks will enable the automatic generation of rewards and reward combinations.

\section*{Acknowledgments}

The authors would like to thank Konstantinos Bousmalis and Patrick Pilarski 
and many others of the DeepMind team for their help and numerous useful discussions and
feedback throughout the preparation of this manuscript. In addition, we would like to particularly thank
Thomas Lampe and Michael Neunert and the robot lab team under the lead of Francesco Nori
for their continuous support with the real robot experiments.

\bibliographystyle{plainnat}
\bibliography{references}

\clearpage
\appendices

\section{Details on the experimental setup}

In the following sections we give a detailed description of the experimental setup for the manipulation and the Ball-in-a-Cup domains.

\subsection{Manipulation setup}

For the experiments in the manipulation setup we use a MuJoCo \citep{todorov2012mujoco} simulation that is well aligned with the real world robot's setup. In both cases, a Rethink Sawyer robotic arm, with a Robotiq 2F-85 parallel gripper as end-effector, faces a 20cm x 20xm basket, that, in the case of the `Lift' and `Lift Any' tasks, contains a single colored block or, in the case of the `Stack` task, two differently colored blocks. The basket is equipped with three cameras, two attached at the front corners of the basket and one attached in the back-left corner. All cameras face the basket's center. The 3D-printed, colored blocks are cubic, with a side length of 5cm.

In simulation, the blocks are initialized at random locations within the 20cm x 20cm workspace at the beginning of each episode. On the real robot an initialization sequence, that randomizes the block positions, is triggered every 20 episodes.

As shown in Table \ref{tab:manipulation_observations}, the agent receives proprioceptive information - the joint positions, joint velocities and joint torques, as well as the wrist sensor's force torque readings. Additionally, the agent is provided with a binary grasp sensor. Further, the agent receives the camera images of all three cameras that are attached to the basket as the only exteroceptive inputs. As shown in Table \ref{tab:manipulation_action_space}, the action space is five dimensional, continuous and consists of the three cartesian translational velocities, the angular velocity of the wrist around the vertical axis and the current speed of the gripper's fingers. The workspace is limited to the 20cm\textsuperscript{3} cube above the table surface. 

To determine a successful lift, we compute the end-effector's tool center point (TCP) using forward kinematics. The `Lift' and `Lift Any' rewards are given in terms of the TCP's z-coordinate $\textrm{TCP}_z$ by
\begin{equation*}
    r_\textrm{lift} = 
    \begin{cases}
      1 & \text{if $\mathrm{TCP}_z > 0.15\textrm{m}$ and the grasp sensor is active,}\\
      0 & \text{otherwise.}
    \end{cases}   
\end{equation*}

\noindent Additionally, on the real robot, we measure external forces at the wrist sensor for safety reasons and the episode is terminated if a threshold of $20\mathrm{N}$ on any of the three principle axes is exceeded for more than three consecutive time steps. If this happens, the episode is terminated with zero reward, enforcing the agent to avoid these failure situations in the future.

\begin{table}
\caption{Observation space of the manipulation setup.}
\label{tab:manipulation_observations}
\vskip 0.15in
\begin{center}
\begin{small}
\begin{tabular}{lcc}
\toprule
Entry & Dimensions & Unit \\
\midrule
Joint Position (Arm) & 7 & rad \\
Joint Velocity (Arm) & 7 & rad/s \\
Joint Torque (Arm) & 7 & Nm \\
Joint Position (Hand) & 1 & rad \\
Joint Velocity (Hand) & 1 & tics/s \\
Force-Torque (Wrist) & 6 & N, Nm \\
Binary Grasp Sensor & 1 & au \\
TCP Pose & 7 & m, au \\
Camera Image (Front Right) & 64 x 64 x 3 & \\
Camera Image (Front Left) & 64 x 64 x 3 & \\
Camera Image (Back Left) & 64 x 64 x 3 & \\
Last Control Command (Joint Velocity) & 8 & rad/s \\
\bottomrule
\end{tabular}
\end{small}
\end{center}
\vskip -0.1in
\end{table}

\begin{table}
\caption{Action space of the manipulation setup.}
\label{tab:manipulation_action_space}
\vskip 0.15in
\begin{center}
\begin{small}
\begin{tabular}{lccc}
\toprule
Entry & Dimensions & Unit & Range \\
\midrule
Translational Velocity (in x, y, z) & 3 & m/s & [-0.07, 0.07] \\
Wrist Rotation Velocity & 1 & rad/s & [-1, 1] \\
Finger speed & 1 & tics/s & [-255, 255] \\
\bottomrule
\end{tabular}
\end{small}
\end{center}
\vskip -0.1in
\end{table}

\begin{table}
\caption{Observation space of the Ball-in-a-Cup setup.}
\label{tab:bic_observations}
\vskip 0.15in
\begin{center}
\begin{small}
\begin{tabular}{lcc}
\toprule
Entry & Dimensions & Unit \\
\midrule
Joint Position (Arm) & 7 & rad \\
Joint Velocity (Arm) & 7 & rad/s \\
Camera Image (Frontal) & 84 x 84 x 3 & \\
Camera Image (Side) & 84 x 84 x 3 & \\
Last Control Command (Joint Velocity) & 4 & rad/s \\
Action Filter State & 4 & rad/s \\
\bottomrule
\end{tabular}
\end{small}
\end{center}
\vskip -0.1in
\end{table}

\begin{table}
\caption{Joint limits of the Ball-in-a-Cup setup.}
\label{tab:bic_limits}
\vskip 0.15in
\begin{center}
\begin{small}
\begin{tabular}{lccccc}
\toprule
 & Unit & J0 & J1 & J5 & J6 \\
\midrule
Position & rad & [-0.4, 0.4] & [0.3, 0.8] & [0.5, 1.34] & [2.6, 4.0] \\
Velocity & rad/s & [-2.0, 2.0] & [-2.0, 2.0] & [-2.0, 2.0] & [-2.0, 2.0] \\
\bottomrule
\end{tabular}
\end{small}
\end{center}
\vskip -0.1in
\end{table}

\subsection{Ball-in-a-Cup}
The Ball-in-a-Cup setup features a Rethink Sawyer robotic arm mounted on a stand, with a custom, 3D-printed cup attachment, which is fixed to the arm's wrist. The cup has a diameter of 20cm and a height of 16cm. A woven net is attached to the cup, to ensure, that the ball is visible even when it is inside the cup. The ball has a diameter of 5cm and is attached to a ball bearing at the robot's wrist with 40cm Kevlar string. The bearing helps to prevent winding of the string -- when the ball goes around the wrist, the string rotates freely via the bearing. The Ball-in-a-cup cell is equipped with two cameras positioned orthogonally -- and both facing the robot -- as well as with a Vicon Vero \citep{ViconWeb} setup, that tracks the cup and the ball and which is used for computing the sparse, external `catch' reward, as well as for triggering resets.

The action space 4 dimensional, continuous and consists of the joint velocities to be commanded for 4 of the 7 joints (J0, J1, J5 and J6), which are sufficient to solve the task. The unused degrees of freedom are commanded to remain at a fixed position throughout the episode. The imposed joint limits are shown in Table \ref{tab:bic_limits} and are chosen as to avoid collisions with the workspace and to keep the ball and cup visible in both cameras at all times. To prevent high frequency oscillations in the commanded velocities, the outputs of the policy network are passed through a low-pass filter with a cutoff frequency of 0.5 Hz.

As shown in Table \ref{tab:bic_observations}, the agent receives proprioceptive information - the joint positions and joint velocities, as well as the camera images of the two cameras attached to the cell. Further, the agent is provided with the action executed at the previous time step as well as with the low-pass filter's internal state.

Due to the string wrapping and tangling around the robot and cup, resets can be challenging. In order to minimize human intervention during training, the robot uses a simple hand-coded reset policy that allows for mostly unattended training. We refer to \citep{schwab19simultaneously} for details on the reset procedure.

We use the 3D pose of the ball and the cup, obtained from the Vicon Vero tracking system, to determine the sparse, external reward, which is only given in case of a successful catch:

\begin{equation*}
    r_\textrm{catch} = 
    \begin{cases}
      1 & \text{if ball in cup,}\\
      0 & \text{otherwise.}
    \end{cases}   
\end{equation*}

\section{Network architecture and Hyperparameters}

\begin{table}[t]
\caption{Hyperparameters for MPO and SAC-X.}
\label{tab:hyperparams}
\vskip 0.15in
\begin{center}
\begin{small}
\begin{tabular}{lc}
\toprule
2D Conv Channels & 128, 64, 64 \\
2D Conv Kernel Sizes & 4x4, 3x3, 3x3 \\
2D Conv Strides & 2, 2, 2 \\
Actor Torso Layer Sizes & 256 \\
Actor Head Layer Sizes & 100, 16 or 8 \\
Critic Torso Layer Sizes & 400 \\
Critic Head Layer Sizes & 300 , 1 \\
Activation Function & elu \\
\midrule
Discount Factor & 0.99 \\
Adam Learning Rate & 2e-4 \\
Replay Buffer Size & 1e6 \\
Target Network Update Period & 500 \\
Batch Size & 64 \\
KL Constraint Epsilon & 1 \\
KL Epsilon Samples & 0.1 \\
KL Epsilon Mean & 1e-3 \\
KL Epsilon Covariance & 1e-5 \\
\bottomrule
\end{tabular}
\end{small}
\end{center}
\vskip -0.1in
\end{table}

In all experiments we use Scheduled Auxiliary Control (SAC-X) \citep{Riedmiller2018_Learning}, either with a uniform random scheduler (SAC-U) or with a learned, Q-table based scheduler (SAC-Q). However, in contrast to the original work, where stochastic value gradients \citep{heess2015learning} are used for optimization, we use Maximum a Posteriori Policy Optimisation (MPO) \citep{abdolmaleki2018maximum} for learning the policy and value function.
Following \citep{Riedmiller2018_Learning}, we implement both the policy and value function with multi-headed (multi-task) neural networks. All pixel inputs are passed through a convolutional neural network with three convolutional layers. The resulting embeddings are subsequently concatenated with the proprioceptive inputs. The torso is shared between the network heads and comprises a single fully-connected layer. The task heads in turn consist of two additional fully-connected layers. The final layer outputs the parameters of a multivariate Gaussian distribution, with the same dimensionality as the action space. The network details and hyperparameter values are shown in Table \ref{tab:hyperparams}.

\end{document}